\documentclass[11pt]{article}

\usepackage[preprint]{acl}
\usepackage{booktabs} % Recommended for professional, well-spaced tables
\usepackage{amsmath}  % Useful for mathematical formulas
\usepackage{times}
\usepackage{latexsym}
\usepackage{booktabs}
\usepackage{multirow}
\usepackage{color}
\usepackage{amssymb}
\usepackage{graphicx}    % 插图（必须）
\usepackage{subcaption} % subfigure / subcaption（推荐）
\usepackage{xcolor}
\usepackage{colortbl}
\usepackage{adjustbox}
\usepackage[T1]{fontenc}
\usepackage[utf8]{inputenc}

\usepackage{microtype}

\usepackage{inconsolata}

\usepackage{graphicx}

\title{Confidence over Time: Confidence Calibration with \\Temporal Logic for Large Language Model Reasoning}
\author{
{\bf Zhenjiang Mao}$^{1}$\thanks{Equal contribution.},
{\bf Anirudhh Venkat}$^{1}$\footnotemark[1],
{\bf Artem Bisliouk}$^{2}$,
{\bf Akshat Kothiyal}$^{1}$ \\
{\bf Sindhura Kumbakonam Subramanian}$^{1}$,
{\bf Saithej Singhu}$^{1}$,
{\bf Ivan Ruchkin}$^{1}$ \\
\\
$^{1}$University of Florida\\
$^{2}$University of Mannheim \\
}

\begin{document}
\maketitle
\begin{abstract}

% Reliable confidence estimation is essential for using large language models (LLMs) in error-sensitive domains. Sadly, existing uncertainty quantification ignores the temporal evolution of confidence in long-form reasoning responses. 
% We study both globally shared STL blocks and an instance-adaptive variant that predicts question-specific parameters via a hypernetwork while preserving structural interpretability. 
Large Language Models (LLMs) increasingly rely on long-form, multi-step reasoning to solve complex tasks such as mathematical problem solving and scientific question answering.
Despite strong performance, existing confidence estimation methods typically reduce an entire reasoning process to a single scalar score, ignoring how confidence evolves throughout the generation.
As a result, these methods are often sensitive to superficial factors such as response length or verbosity, and struggle to distinguish correct reasoning from confidently stated errors.
We propose to characterize the stepwise confidence signal using Signal Temporal Logic (STL). Using a discriminative STL mining procedure, we discover temporal formulas that distinguish confidence signals of correct and incorrect responses.
Our analysis found that the STL patterns generalize across tasks, and numeric parameters exhibit sensitivity to individual questions.  Based on these insights, we develop a confidence estimation approach that informs STL blocks with parameter hypernetworks. 
Experiments on multiple reasoning tasks show our confidence scores are more calibrated than the
%  STL-based confidence estimation consistently improves calibration, achieving lower Expected Calibration Error, Brier Score, and Negative Log-Likelihood compared to main-stream
baselines.

% 8 pages 

\end{abstract}

\section{Introduction}

\looseness=-1
Large Language Models (LLMs) have demonstrated strong performance on long-form, multi-step reasoning involved in mathematical problem solving, scientific explanation, and legal analysis~\cite{wei2022chain,wang2022self,yao2023tree}. Despite their fluency, LLMs remain prone to factually incorrect or logically flawed responses that are expressed with high apparent certainty~\cite{kadavath2022language}.
% Recent work has also explored enforcing formal temporal constraints during LLM generation to guarantee compliance with safety specifications, for example, by constraining token-level actions using logic-based verification techniques~\citep{kamath2025enforcing}.
Recent work has also explored enforcing formal temporal constraints during LLM generation to improve compliance with safety specifications, for example via logic-based verification techniques~\citep{kamath2025enforcing}.
% This phenomenon, commonly referred to as hallucination~\cite{ji2023survey}, poses a serious risk in error-sensitive domains and motivates the need for \textit{reliable confidence estimation} for individual model outputs. Specifically, we estimate a sample-level probability that a given generated response is correct conditioned on the input --- and evaluate how well this estimate is calibrated against empirical correctness.
Specifically, we estimate a sample-level probability that a generated response is correct conditioned on the input, and evaluate its calibration against empirical correctness.

% Our goal is to estimate the probability that a given generated response is correct, conditioned on the input prompt, and to assess how well this estimate is calibrated with respect to empirical correctness.

\begin{figure*}[t]
  \includegraphics[width=2\columnwidth]{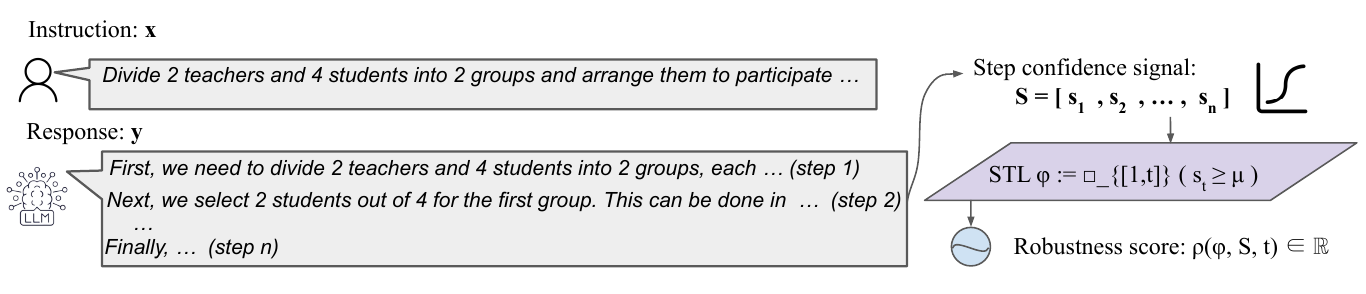}
%   \caption{Illustration of stepwise confidence modeling with STL.
% An LLM-generated reasoning response is segmented into steps, and token-level
% probabilities are aggregated into a stepwise confidence signal
% $\mathbf{S}$.
% An STL formula (e.g., $\varphi := \Box_{[1,t]}(s_t \ge \mu)$) is evaluated on
% $\mathbf{S}$ to capture temporal confidence patterns.
% The resulting robustness score $\rho(\varphi,\mathbf{S},t)$
% quantifies how strongly the confidence trajectory satisfies the specified temporal constraint.}
\caption{Overview of stepwise confidence modeling with STL.
Token-level probabilities are aggregated into a stepwise confidence signal $\mathbf{S}$, on which STL formulas are evaluated to capture temporal confidence patterns via robustness scores.}
% temporal constraint, where $t$ denotes the evaluation time step.}
\vspace{-0.5cm}
  \label{fig:stl}
\end{figure*}

Most existing approaches to confidence or uncertainty quantification for LLMs reduce the generation process to a single scalar score. Representative methods include aggregating token-level probabilities, computing entropy-based statistics, or estimating confidence through output consistency across multiple samples~\cite{ott2018analyzing}. While effective in certain settings, these share a \textit{common limitation:} they collapse a sequential reasoning process into a scalar, discarding information about how confidence evolves over time.
% they collapse a sequential reasoning process into a scalar without regard for the temporal structure of the reasoning process, thereby discarding information about how confidence evolves throughout a long reasoning response. 
This makes the confidence depend heavily on how the response is verbalized (length, verbosity, and stylistic fillers), since token-level aggregates can be dominated by high-probability but low-semantic-content tokens.

\looseness=-1
% We posit that confidence in long-form LLM reasoning should be viewed not as a static scalar, but as a \textit{temporal signal}~\cite{dong2018confidence,ribeiro2020beyond} that carries information about correctness. During generation, an LLM makes a sequence of stochastic decisions, and the resulting confidence trajectory exhibits characteristic patterns depending on whether the underlying reasoning is valid or flawed. 
% % For example, correct responses may maintain consistently high confidence across key reasoning segments, whereas incorrect responses may display instability, abrupt drops, or oscillations. 
% For example, correct responses tend to maintain stable confidence across reasoning steps, whereas incorrect responses often exhibit instability or oscillation.
% % Capturing such temporal structure requires moving beyond token aggregation toward taking confidence dynamics into account.

% This paper proposes to derive confidence as a stepwise signal from segmented reasoning responses and characterize its temporal evolution using \textit{Signal Temporal Logic (STL)}~\citep{maler2004monitoring}. 
% STL provides a formal language for specifying temporal patterns~\cite{linard2020active,nicoletti2024mining} over real-valued signals and yields a quantitative robustness score indicating how well a signal satisfies a temporal specification (Figure~\ref{fig:stl}).

We posit that confidence in long-form LLM reasoning should be viewed not as a static scalar, but as a \textit{temporal signal}~\cite{dong2018confidence,ribeiro2020beyond} that reflects the evolution of model certainty over reasoning steps.
To capture such temporal patterns, we derive stepwise confidence signals from segmented reasoning responses and characterize them using \textit{Signal Temporal Logic (STL)}~\citep{maler2004monitoring}, which provides a formal language for specifying temporal constraints over real-valued signals~\cite{linard2020active,nicoletti2024mining}.
STL yields a quantitative robustness score that indicates how well a confidence trajectory satisfies a given temporal specification (Figure~\ref{fig:stl}).
Guided by STL robustness, we use a discriminative STL mining procedure to automatically discover temporal formulas that best distinguish correct from incorrect LLM responses.
% STL provides a formal language for specifying temporal patterns~\cite{linard2020active,nicoletti2024mining} over real-valued signals and, through its quantitative semantics, yields a \textit{robustness score} that measures the degree to which a signal satisfies a given temporal specification, as shown in Figure~\ref{fig:stl}. 
 %The resulting robustness scores provide an interpretable \emph{confidence signal}, which can be mapped to calibrated probability estimates via a lightweight monotonic transformation.

% Beyond predictive performance, 
% We systematically investigate the structure and variability of temporal confidence patterns. 
% We systematically analyze the structure and variability of temporal confidence patterns to inform model design.
% Specifically, we ask whether the \emph{structure} of temporal patterns is shared across datasets and task types, or whether different reasoning tasks require fundamentally different specifications.
% Further, within a fixed task setting, we examine whether a single set of temporal \emph{parameters} suffices across instances, or whether question-level adaptation is necessary~\cite{ha2016hypernetworks,garnelo2018conditional} to capture variability in confidence dynamics.
% These investigations guide our design of how much question-level adaptation is required for confidence estimation. %at the question level.

We analyze the structure and variability of temporal confidence patterns across tasks and instances, and use these insights to guide the design of our confidence estimation framework.
Based on these insights, we design an interpretable confidence estimation framework that separates temporal structure from parameter instantiation.
We fix STL structures to preserve interpretability, while enabling question-level adaptation of their continuous parameters via a hypernetwork~\cite{ha2016hypernetworks,garnelo2018conditional}.

% To exploit temporal modeling for confidence calibration, we design a modular architecture based on \emph{STL blocks}.
% Each STL block wraps around a fixed temporal logic formula to map a stepwise confidence trajectory to an interpretable confidence score.
% While reusing an STL block would be beneficial when its temporal pattern generalizes, different input questions may vary in reasoning difficulty and the emergent confidence dynamics.
% To capture such question-level variability without sacrificing interpretability, we further propose an instance-adaptive design in which the STL structure is fixed, but its continuous parameters are predicted on a per-question basis via a \emph{hypernetwork}~\cite{ha2017hypernetworks}.
% This design explicitly separates temporal structure from parameter instantiation.

\looseness=-1
Our contributions are as follows: 
(1) We introduce discriminative STL mining for modeling stepwise confidence signals in long-form LLM reasoning;
(2) We show that effective temporal confidence structures exhibit consistent patterns across datasets and task types, suggesting reusable temporal signatures, particularly for incorrect reasoning;
(3) Within a fixed task setting, we show that while a single STL structure can generalize across questions, its optimal parameters can exhibit significant question-level variability;
(4) We propose an interpretable confidence estimation framework that combines STL robustness with a hypernetwork-based parameterization, improving standard calibration metrics while preserving temporal structure.

\section{Background and Problem}
\label{sec:problem}

\subsection{Probability in LLMs}

We consider a family of language tasks indexed by task type $\mathcal{T}$.
Given an input prompt $\mathbf{x}$, a Large Language Model (LLM) $\mathcal{M}$
generates a response ${\mathbf{y}} = [{y}_1, \dots, {y}_m]$ in an
autoregressive manner.
At each decoding step, the model assigns a probability to the generated token
conditioned on the input and preceding tokens.
We denote by
$c_{\mathcal{T}}(\mathbf{x}, {\mathbf{y}}) \in \{0,1\}$
a binary correctness predicate indicating whether the generated response
${\mathbf{y}}$ is correct for input $\mathbf{x}$ under task type $\mathcal{T}$,
as determined by task-specific evaluation protocols.
The goal of confidence estimation is to produce a scalar score reflecting the
likelihood that a generated response is correct.
A common baseline is to use the joint generation probability
$P({\mathbf{y}} \mid \mathbf{x})$ as confidence.
However, this quantity suffers from severe length bias, assigning smaller
values to longer responses regardless of correctness~\cite{li2025language},
making it unsuitable for long-form reasoning.

\noindent
\textbf{Problem Formulation.}
Our objective is to estimate a confidence score
$\hat{P} \in [0,1]$ that approximates the posterior probability
$P(c_{\mathcal{T}}(\mathbf{x}, {\mathbf{y}})=1 \mid \mathbf{x})$.
We assume access to labeled input--output pairs
$\{(\mathbf{x}_i, {\mathbf{y}}_i, c_i)\}_{i=1}^N$
and learn a confidence model $\mathcal{S}$ mapping
$(\mathbf{x}, {\mathbf{y}})$ to $\hat{P} = \mathcal{S}(\mathbf{x}, {\mathbf{y}})$.

\subsection{Evaluation Metrics}
We evaluate confidence estimation quality using standard calibration metrics, primarily
\emph{Expected Calibration Error (ECE)}~\cite{guo2017calibration} and the
\emph{Brier Score (BS)}~\cite{glenn1950verification}, which quantify the alignment between predicted confidence and empirical correctness.
Formal definitions are provided in the appendix.

% \noindent
% \textbf{Negative Log-Likelihood (NLL):} NLL penalizes the model heavily for assigning low probability to correct answers (or high probability to incorrect ones), defined as:
% \begin{align}
%     \text{NLL} &= - \frac{1}{N} \sum_{i=1}^{N} \text{BCE}(c_i, {P}_i) \label{eq:NLL} \\
%     \begin{split}
%     \text{BCE}(c_i, {P}_i) &= \left[ c_i \log({P}_i) \right. \\
%     &\quad \left. + (1 - c_i) \log(1 - {P}_i) \right] \label{eq:BCE}
%     \end{split}
% \end{align}
    % \begin{equation}
    %     \text{NLL} = - \frac{1}{N} \sum_{i=1}^{N} \left[ c_i \log({P}_i) + (1 - c_i) \log(1 - {P}_i) \right].
    % \end{equation}

\subsection{Stepwise Confidence Signal and STL}

We represent a generated response ${\mathbf{y}}$ 
as a sequence of $n$ consecutive segments,
% as a sequence of
% $n$ consecutive segments, denoted by
% $\mathcal{U} = \{u_1, u_2, \dots, u_n\}$,
% where each segment $u_j$ corresponds to a contiguous subsequence of tokens.
where each segment corresponds to a contiguous subsequence of tokens.
This segmentation induces a stepwise representation of the generation process
and serves as the temporal axis for subsequent confidence modeling.
In practice, such a decomposition may follow linguistic boundaries
(e.g., sentences) or semantic reasoning steps
(e.g., those produced by Chain-of-Thought prompting).
Our formulation requires consistent segmentation within each task --- but does not assume any specific method, 
which is treated as a design choice.
% which is treated as a design choice and is
% discussed in the context of our approach and experiments. 
%While different segmentation strategies may lead to different signal resolutions, our formulation does not assume a specific segmentation scheme and only requires a consistent decomposition within each task setting.
% Formally, each segment $u_j$ is a contiguous subsequence of tokens:
% \begin{equation}
% u_j = [ y_{t_j},  y_{t_j+1}, \dots,  y_{t_j+L_j-1}],
% \end{equation}
% where $t_j$ and $L_j$ are the respective start and length of the $j$-th segment.
% Let $r_j$ consist of a subsequence of tokens $\{y_{t}, y_{t+1}, \dots, y_{t+L_j-1}\}$, where $L_j$ is the length of the $j$-th step. 

We define the \textit{stepwise confidence signal} $s_j$ as the aggregated confidence associated with segment $u_j$.
Specifically, we compute $s_j$ as the arithmetic mean of the probabilities of its constituent tokens:
\begin{equation}
s_j = \frac{1}{L_j} \sum_{k=0}^{L_j-1}
P( y_{t_j+k} \mid {\mathbf{y}}_{<t_j+k}, \mathbf{x}).
\end{equation}

This process transforms the token-level probability sequence into a coarser-grained confidence signal $\mathbf{s} = [s_1, s_2, \dots, s_n]$, which serves as the foundation for logic-based modeling.

\noindent
\textbf{STL Syntax.}
Signal Temporal Logic (STL) provides a formalism for specifying temporal constraints over real-valued signals~\cite{maler2004monitoring}.
In this work, we use STL to describe temporal patterns over the stepwise confidence signal $\mathbf{S} = [s_1, \dots, s_n]$.

An STL formula consists of a fixed logical structure $\varphi$ together with continuous parameters $\boldsymbol{\theta}$, such as predicate thresholds and temporal bounds, and is denoted by $\varphi_{\boldsymbol{\theta}}$.
The basic building block is a predicate $\mu$, typically of the form $\mu \equiv s_t \geq c$, where larger values indicate higher confidence.

\noindent
\textbf{STL Robustness.}
STL formulas are evaluated under quantitative robustness semantics, yielding a score that summarizes how well a stepwise confidence trajectory satisfies a temporal pattern.

\section{Related Work}

\subsection{Uncertainty Quantification for LLMs}

% In this work, we focus on \emph{confidence estimation} at the response level,
% i.e., predicting the probability that a specific generated response is correct,
% rather than explicitly modeling or decomposing aleatoric or epistemic uncertainty.

We study response-level \emph{confidence estimation}: predicting the probability that a specific generated response is correct, rather than decomposing aleatoric/epistemic uncertainty.
While UQ is well-studied in MT~\citep{ott2018analyzing, wang-etal-2021-beyond-glass} and classification~\citep{beck-etal-2016-exploring}, free-form LLM generation is harder due to its effectively unbounded output space.
% While uncertainty estimation has been extensively studied in conventional NLP tasks such as machine translation \citep{ott2018analyzing, wang-etal-2021-beyond-glass} and classification \citep{beck-etal-2016-exploring}, quantifying uncertainty in free-form LLM generation 
% % presents unique challenges
% % due to the flexible and effectively unbounded \emph{output space} of natural
% % language responses.
% presents unique challenges due to the effectively unbounded output space of free-form generation.
Existing approaches can be broadly categorized into \emph{four lines of research}: logit-based, verbalized uncertainty, consistency-based, and internal-state-based.

\noindent
\textbf{Logit-based:} 
% Early approaches to autoregressive UQ rely on the model's output probabilities. Standard metrics include computing cumulative \textit{token-wise} probabilities, predictive entropy, or max probability \citep{kadavath2022language, jiang2021can}. 
Early autoregressive UQ approaches rely on output probabilities, including cumulative token-wise likelihoods, predictive entropy, and max probability \citep{kadavath2022language, jiang2021can}.
% To mitigate the bias introduced by sentence length, length-normalized variants have also been proposed \citep{malinin2021uncertainty}. 
Length-normalized variants have been proposed to mitigate sentence-length bias \citep{malinin2021uncertainty}.
However, for long-form reasoning they conflate semantic and auxiliary tokens, often causing miscalibration.

% However, these struggle with long reasoning sequences, 
% where semantic content is conflated with auxiliary tokens, often leading to miscalibration in long reasoning sequences.
% where semantic content and auxiliary tokens (e.g., transition words) are conflated in joint probability estimations, often leading to miscalibration.

\noindent
\textbf{Verbalized Uncertainty:}
% Another category explores the capability of LLMs to explicitly express their own uncertainty in natural language. Research demonstrates that models can be prompted or fine-tuned to output calibrated confidence statements \citep{lin2022teachingmodelsexpressuncertainty, tian2023justaskcalibrationstrategies}. 
% Another category explores verbalized uncertainty, where LLMs are prompted or fine-tuned to explicitly express confidence in natural language \citep{lin2022teachingmodelsexpressuncertainty, tian2023justaskcalibrationstrategies}.
% Representative approaches include prompt- and sampling-based calibration \citep{xiong2024can} and targeted fine-tuning strategies \citep{amayuelas2024knowledge}.
Verbalized uncertainty prompts or fine-tunes LLMs to express confidence in natural language~\citep{lin2022teachingmodelsexpressuncertainty, tian2023justaskcalibrationstrategies}, including prompt/sampling schemes~\citep{xiong2024can} and targeted fine-tuning~\citep{amayuelas2024knowledge}; these often require additional training or domain adaptation~\citep{kadavath2022language, manakul2023selfcheckgpt}.
Although promising, these methods often require additional training (e.g., RLHF or fine-tuning) or domain-specific adaptation to perform well \citep{kadavath2022language, manakul2023selfcheckgpt}.
% For example, ~\citealp{xiong2024can} proposed a framework featuring tailored prompts and sampling to reduce overconfidence, while ~\citealp{amayuelas2024knowledge} improved calibration through targeted fine-tuning. Although promising, these methods often require additional training (RLHF or fine-tuning) or domain-specific adaptations to perform well \citep{kadavath2022language, manakul2023selfcheckgpt}.

% \paragraph{ and Semantic Clustering.}

\noindent
 \textbf{Consistency-based:}
% To address the intractable reasoning space and the issue of ``semantic equivalence,''
% a prominent line of work approximates uncertainty through a semantic space
% constructed from multiple sampled outputs.
To address the intractable reasoning space and semantic equivalence, a prominent line of work estimates uncertainty from multiple sampled outputs in a semantic space.
\emph{Self-Consistency}~\citep{wang2023selfconsistency} estimates confidence from agreement among diverse generations, while \textit{Semantic Entropy}~\citep{kuhn2023semantic} formalizes this idea via semantic clustering and entropy-based scoring.
% \emph{Self-Consistency}~\citep{wang2023selfconsistency} is a representative paradigm in this category, which posits that diverse yet semantically consistent generations
% indicate high confidence.
% \textit{Semantic Entropy} (SE)~\cite{kuhn2023semantic} further formalizes this idea
% by clustering similar responses and measuring uncertainty via normalized cumulative
% log probabilities.
%  This approach has been extended to black-box models using co-occurrence entailment matrices \citep{farquhar2024detecting} and combined with self-reflective evaluation for enhanced robustness \citep{kirchhof2025self}.
% Other variations include using paraphrased prompts to gather multiple outputs \citep{li2025principled} or appending clarification sequences to differentiate between aleatoric and epistemic uncertainty \citep{10.5555/3692070.3692835}.
% Extensions include black-box formulations using entailment matrices \citep{farquhar2024detecting}, self-reflective evaluation \citep{kirchhof2025self}, paraphrased prompting \citep{li2025principled}, and clarification-based strategies for disentangling uncertainty types \citep{10.5555/3692070.3692835}.
Extensions use entailment-based black-box clustering~\citep{farquhar2024detecting}, self-reflection~\citep{kirchhof2025self}, paraphrased prompting~\citep{li2025principled}, and clarification sequences~\citep{10.5555/3692070.3692835}.
Despite their effectiveness, these methods typically require multiple model runs and often rely on additional components such as external encoder models, introducing non-trivial inference overhead.
% mirror the computational cost of ensemble techniques, requiring multiple model runs and often relying on external encoder models for clustering, which increases overhead.

% \paragraph{-State and Token Attribution.}

\noindent
\textbf{Internals-based:}
% A fourth line of work estimates uncertainty from internal model states or token-level contributions. Representative approaches include Shifting Attention to Relevance (SAR)~\citep{duan-etal-2024-shifting}, which quantifies token importance via token removal.
% A fourth line of research aims to open the black box by utilizing internal states (e.g., hidden activations, attention heads) or evaluating token-level contributions to the final semantic meaning. 
% For instance, Shifting Attention to Relevance (SAR) was proposed by~\citealp{duan-etal-2024-shifting}, which quantifies token importance by measuring the negative similarity between the complete sequence and a version with the token removed. 
% While these ``white-box'' or internal-based methods \citep{beigi-etal-2024-internalinspector, azaria-mitchell-2023-internal} offer granular insights, they often necessitate computationally expensive procedures 
% While such internal-based methods~\citep{beigi-etal-2024-internalinspector, azaria-mitchell-2023-internal} provide granular insights, they often require computationally expensive procedures 
% (e.g., factorial complexity in SAR) or rely on large in-domain validation sets to identify informative attention heads, limiting their practicality in real-world contexts.
Internals-based methods estimate uncertainty from hidden states or token attribution, e.g., SAR~\citep{duan-etal-2024-shifting}; related approaches exploit model internals~\citep{beigi-etal-2024-internalinspector, azaria-mitchell-2023-internal} but can be expensive (e.g., factorial SAR) or require large in-domain validation sets.
% \paragraph{Verbalized Uncertainty and Self-Evaluation.}
% Despite this progress, LLMs remain overconfident, particularly in zero-shot or out-of-domain settings~\citep{liu2025uncertainty, desai-durrett-2020-calibration}, which is problematic in domains such as education and math problem solving.
% Moreover, most methods are evaluated on general NLP tasks, leaving structured reasoning scenarios under-explored.
Despite progress, LLMs remain overconfident in zero-shot and out-of-domain settings~\citep{liu2025uncertainty, desai-durrett-2020-calibration}, and structured reasoning remains comparatively under-explored.
% Despite the progress, LLMs continue to suffer from overconfidence, particularly in zero-shot or out-of-domain tasks \citep{liu2025uncertainty, desai-durrett-2020-calibration}. This issue is critical in sensitive domains such as education and math problem solving, where reliable uncertainty estimates are essential for helping students assess model-generated reasoning. Most existing methods have been evaluated on general NLP tasks, leaving structured reasoning scenarios relatively under-explored.

% \subsection{Confidence and Calibration}

\begin{figure*}[t]
  \includegraphics[width=2\columnwidth]{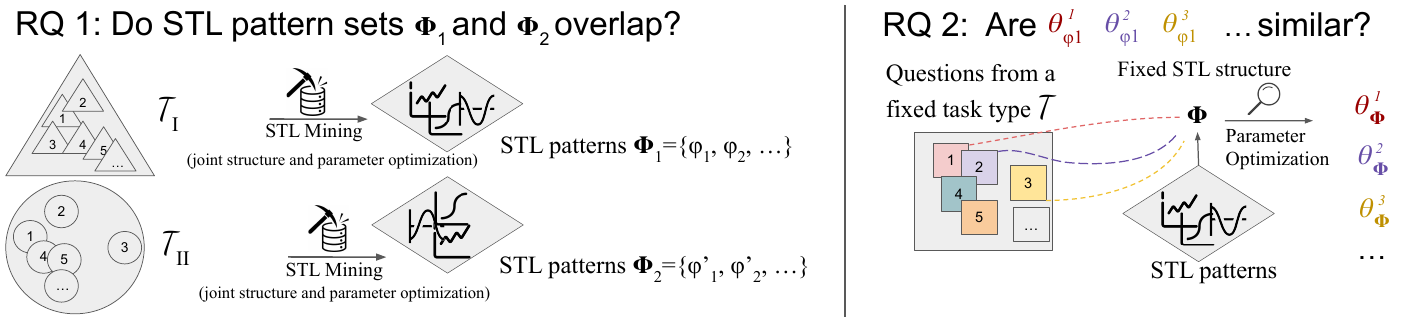}
  \caption{Conceptual illustration of Research Question~1 (RQ1) and Research Question~2 (RQ2).
RQ1 examines whether the STL structures mined from different task types are shared or task-specific.
RQ2 studies, under a fixed STL structure, whether the associated parameters $\theta$ learned during optimization generalize across questions.}
\vspace{-4mm}
%or require question-level adaptation.}
  \label{fig:rqs}
\end{figure*}

% Temporal logic, originally developed for formal verification of hardware and software systems \citep{pnueli1977temporal}, provides a rigorous language for specifying system behaviors over time. 
% While Linear Temporal Logic (LTL) deals with discrete states, \textit{Signal Temporal Logic} (STL) \citep{maler2004monitoring} was introduced to express properties of real-valued, continuous-time signals (e.g., physical sensors or, in our case, probability trajectories). 
% A defining feature of STL is its \textit{quantitative semantics} (or robustness degree), which maps a trace and a formula to a continuous scalar value $\rho \in \mathbb{R}$, rather than a Boolean outcome. 
% This robustness measures \textit{how strongly} a property is satisfied or violated, enabling synergies of logic and optimization. 

\subsection{Temporal Logics}

% Temporal logic, originally developed for formal verification of hardware and software systems~\citep{pnueli1977temporal}, provides a principled framework for specifying temporal patterns over signals.
% \textit{Signal Temporal Logic} (STL)~\citep{maler2004monitoring} extends this framework to real-valued signals and, through its quantitative semantics, assigns a robustness score that measures how strongly a temporal specification is satisfied.

Temporal logic, originally developed for formal verification~\citep{pnueli1977temporal}, provides a principled framework for specifying temporal patterns over signals, and \textit{Signal Temporal Logic} (STL)~\citep{maler2004monitoring} extends it to real-valued signals with quantitative robustness semantics. In this work, we apply STL to discrete-time confidence trajectories derived from stepwise LLM reasoning, using robustness to summarize temporal confidence patterns.
% In this work, we apply STL to \emph{discrete-time} confidence trajectories derived from stepwise LLM reasoning, using robustness as a quantitative summary of temporal confidence patterns.
The quantitative nature of STL has enabled its integration into neuro-symbolic learning~\citep{garcez2019neural}, including differentiable constraint enforcement in robotics and time-series models~\citep{raman2014model, 10.1177/02783649221082115}, as well as logic structure mining methods such as TeLEx~\citep{jha2019telex}.

 In NLP, temporal logic has mainly been used for controlled generation~\citep{lu-etal-2021-neurologic} and consistency/fact-checking~\citep{durrett-klein-2014-joint}, focusing on text semantics.
STL has also been used to analyze confidence trajectories post hoc~\citep{mao-etal-2025-temporalizing}; in contrast, we learn STL structures and parameters as \emph{discriminative} confidence estimators under calibration objectives, with question-adaptive parameters via hypernetworks.

% Preliminary prior work has explored the use of STL to analyze confidence trajectories in stepwise reasoning, primarily as a post-hoc diagnostic tool~\cite{mao-etal-2025-temporalizing}. 
% While this line of work is also data-driven in nature, it treats temporal logic patterns as descriptive analyses rather than predictive components.
% In contrast, our work formulates STL-based temporal patterns as \emph{learnable and discriminative confidence estimators}, where both the logical structure and its parameters are optimized under calibration-oriented objectives.
% Moreover, we investigate question-level parameter adaptation via hypernetworks, enabling STL structures to be reused while remaining sensitive to instance-specific confidence dynamics.

%Our work leverages STL not as a constraint for generation, but as a diagnostic tool to characterize the \textit{temporal shape} of uncertainty.

% \begin{figure}[t]
%   \includegraphics[width=\columnwidth]{latex/CoT.pdf}
%   \caption{A figure with a caption that runs for more than one line.
%     Example image is usually available through the \texttt{mwe} package
%     without even mentioning it in the preamble.}
%   \label{fig:q}
% \end{figure}

 % Research Questions;of Mining STL Patterns  and Searching Parameters eneralizability Analysis of STL Mining
% \subsection{Overview}
%Overview: Confidence Estimation via STL
% While standard confidence metrics aggregate token probabilities into a single scalar (e.g., via averaging), they often overlook the \textit{temporal dynamics} of the generation process. 

\section{Temporal Confidence Patterns   }\label{sec:rq}

In this section, we study whether stepwise confidence trajectories exhibit
distinct temporal ``confidence signatures'' for correct versus incorrect
reasoning.
% To this end, we first introduce a discriminative STL mining procedure that
% automatically discovers STL formulas whose robustness scores separate correct
% and incorrect reasoning traces, yielding a formula set
% $\Phi = \Phi_{\text{pos}} \cup \Phi_{\text{neg}}$.
We introduce a discriminative STL mining procedure that discovers STL formulas whose robustness separates correct and incorrect reasoning traces, yielding a formula set $\Phi = \Phi_{\text{pos}} \cup \Phi_{\text{neg}}$.
We then use the mined formulas to investigate structural generalization across
tasks (RQ1) and parameter sensitivity at the question level (RQ2).

\subsection{Discriminative STL Mining}

% To discover interpretable temporal patterns that characterize LLM confidence, we rely on automated formula mining adapted from \textsc{TeLEx} (Template-based Learning of Executives)~\cite{jha2019telex}. 
% Unlike in standard STL mining, which fits a formula to a single set of traces, our goal is discriminative mining: finding formulas that maximally distinguish between correct responses (i.e., responses satisfying $c({\mathbf{y}}, \mathbf{x})=1$)
%  and incorrect ones.

To discover interpretable temporal patterns that characterize LLM confidence, we adopt a discriminative STL mining procedure adapted from \textsc{TeLEx}~\citep{jha2019telex}, aimed at separating correct and incorrect reasoning traces. 
% The mining process proceeds through three phases, as shown below.

% , \textit{Base Construction}, \textit{Signal Augmentation}, and \textit{Structural Composition}, integrated with a classification-based objective.
% Our mining method proceeds through three sequential phases:
% \textit{Base Construction}, \textit{Signal Augmentation}, and
% \textit{Structural Composition}, integrated with a differentiable
% confidence mapping and a classification-based objective.

% Overview of the proposed confidence quantification pipeline with STL blocks.
% The training phase consists of discriminative STL mining and hypernetwork

% % is performed on training data to discover a set of temporal STL formulas $\Phi$ and their associated parameter spaces.

% In Stage~2, the mined formulas are instantiated as STL blocks with learnable parameters.
% At test time, the learned STL structures are fixed and applied to stepwise confidence signals to produce a scalar confidence score.
% In Stage~2, the mined formulas are instantiated as differentiable STL blocks, and a hypernetwork is trained to predict instance-specific STL parameters conditioned on input features.
% At test time, the learned STL structures are fixed, and the trained hypernetwork generates parameters for the STL blocks, which are then applied to stepwise confidence signals to produce a scalar confidence score.

% \subsubsection{Incremental Structure Learning}

\noindent
\textbf{Incremental Structure Learning.}
Given a dataset of stepwise confidence signals 
$\mathcal{D} = \{ (\mathbf{S}_i, c_i) \}_{i=1}^N$, 
where each stepwise confidence signal $\mathbf{S} = [s_1, \dots, s_n]$ derived from an LLM response, we consider a predefined and shared set of atomic predicates $\mathcal{P}$ over the signal, each in the form of a simple inequality
(e.g., $s_t \geq \mu$ or $s_t \leq \mu$), where $\mu$ is a learnable threshold parameter.
The mining process then proceeds incrementally over this fixed predicate space:

\noindent
\textbf{Step 1: Base Template Instantiation.}
Given the stepwise confidence signal $\mathbf{S} = [s_1,\dots,s_n]$ and atomic predicates $\mathcal{P}$, 
 we initialize the search from a predefined set of primitive STL templates $\mathcal{T}_{\text{base}}$.
% we initialize the search with a predefined set of primitive STL templates
% $\mathcal{T}_{\text{base}}$, composed of temporal operators applied to
% atomic predicates. 
Representative examples include $\Box_{[a,b]}(s_t \geq \mu)$ and
$\Diamond_{[a,b]}(s_t \geq \mu)$, and for their full list, see
Appendix~\ref{app:templates}. For each template, we fit its parameters to maximize discriminative power between correct and incorrect responses, using a classification-based objective. This step yields a set of parameterized base templates $\{(\varphi_k, \theta_k)\}$ instantiated from predefined structures.

% To learn nested formulas (e.g., $\Box(s_t \geq 0.5 \implies \Diamond (s_k \geq 0.9))$)  without complex parsing trees,   \textit{signal augmentation} is employed.

\noindent
\textbf{Step 2: Robustness-Based Signal Augmentation.}
To enable the construction of nested STL formulas without explicit enumeration, we adopt a robustness-based signal augmentation strategy (\textit{signal lifting})~\cite{jha2019telex,donze2010robust,kong2014temporal}, which uses robustness signals of mined templates as additional features for subsequent formula composition.

\noindent
\textbf{Step 3: Structural Composition.}
We construct more expressive STL formulas by composing discriminative base templates through temporal nesting and Boolean combinations, enabling the representation of diverse temporal confidence patterns.

\noindent
\textbf{Dual-Class Template Mining.}
Using the candidate STL formulas generated through Steps~1--3, we perform dual-class discriminative mining to identify two complementary categories of temporal confidence patterns.

\noindent
\textit{Correct Patterns ($\Phi_{\text{pos}}$):}
These formulas characterize temporal confidence signatures of correct reasoning, assigning higher scores to correct responses than to incorrect ones.

\noindent
\textit{Incorrect Patterns ($\Phi_{\text{neg}}$):}
These formulas capture confidence patterns associated with hallucinations or reasoning failures, where stronger satisfaction indicates a higher likelihood of incorrectness.

These dual sets of patterns operationalize the hypothesis that correct and incorrect reasoning traces exhibit distinct temporal confidence signatures.

Having performed STL mining, we study the generalizability of the mined patterns from two complementary analytical perspectives. Figure~\ref{fig:rqs} provides a schematic overview of the two research questions, illustrating structural generalization across task types (RQ1) and parameter sensitivity under a fixed formula (RQ2).

\begin{figure}[t]
  \centering
  \includegraphics[width=\columnwidth]{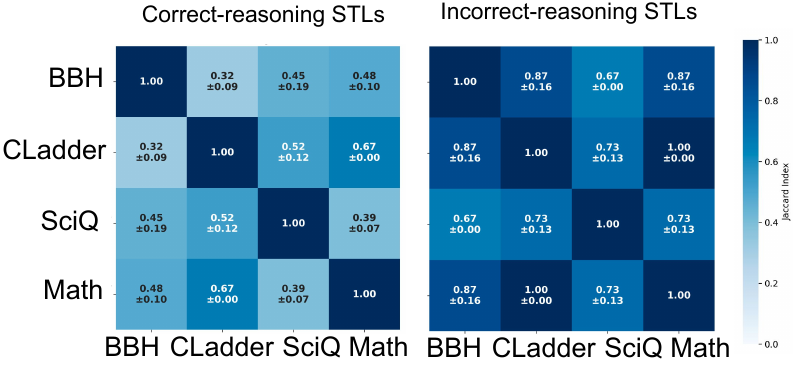}
  \caption{
%   Pairwise similarity of mined STL formulas across four reasoning benchmarks.
% Left: similarity of STL patterns characterizing correct reasoning.
% Right: similarity of STL patterns characterizing incorrect reasoning.
% Similarity is measured using the Jaccard index over mined formula sets and reported as mean $\pm$ standard deviation across folds.
% STL patterns associated with incorrect reasoning exhibit consistently higher cross-task similarity than those associated with correct reasoning, indicating stronger structural reuse across datasets.
Pairwise similarity of mined STL formulas across four reasoning benchmarks.
Left: similarity of STL patterns characterizing correct reasoning.
Right: similarity of STL patterns characterizing incorrect reasoning.
Similarity is measured using the Jaccard index over mined formula sets.
}
\vspace{-5mm}
\label{fig:rq1_overview}
\end{figure}

\subsection{RQ1: Generalizable Structure over Tasks}
\textit{Do effective structural STL patterns generalize across different task types, or are they domain-specific?}
This question examines whether the discriminative STL structures governing temporal confidence are shared across task types or remain task-specific.
We assess this by mining STL formulas on different datasets and measuring the structural overlap of the discovered patterns across tasks.

We empirically observe a clear asymmetry in the structural generalization of temporal confidence patterns.
Across multiple reasoning benchmarks, STL formulas associated with incorrect reasoning exhibit consistently high structural similarity across tasks, whereas those associated with correct reasoning show substantially lower overlap.
This trend holds across backbone models and persists at the subtask level within BBH (see Fig.~\ref{fig:rq1_overview} and Appendix).
These findings indicate that failure-related confidence dynamics are structurally stable and largely reusable across tasks, while successful reasoning trajectories remain task-dependent.
This asymmetry motivates treating $\Phi_{\text{neg}}$ as transferable structural templates, while allowing $\Phi_{\text{pos}}$ to adapt to task-specific characteristics.

\begin{figure}[t]
  \centering
  \includegraphics[width=\columnwidth]{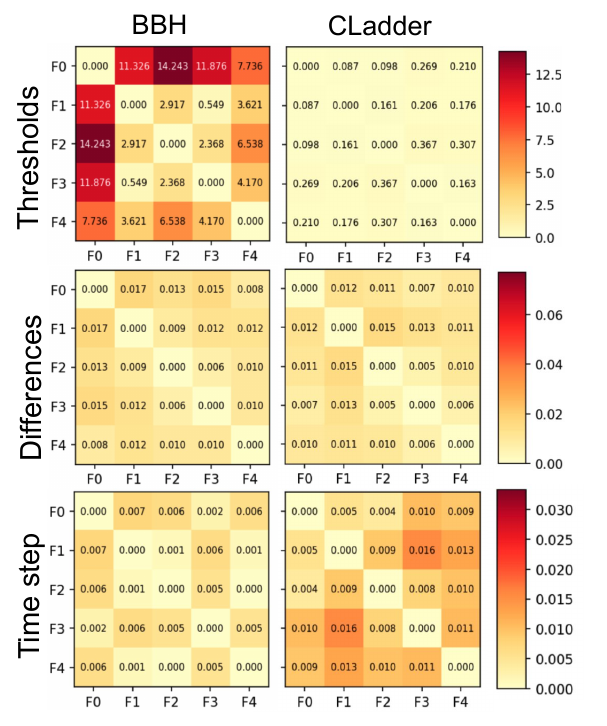}
\caption{
RQ2: Pairwise differences of optimized STL parameters across individual questions under a fixed STL structure.
Parameters are grouped into three categories: predicate thresholds, difference-related parameters, and time-related parameters.
Results are shown for BBH and CLadder using Qwen3-8B.
}
% \caption{
% RQ2: Pairwise differences of optimized STL parameters across individual questions under a fixed STL structure.
% Parameters are grouped into three categories: predicate thresholds (top), difference-related parameters (middle), and time-step-related parameters (bottom).
% Results are shown for two representative datasets (BBH and CLadder) using Qwen3-8B as the backbone model.
% While some parameter categories remain relatively stable across questions, others exhibit substantial variability, indicating heterogeneous parameter sensitivity at the instance level.
% }
\vspace{-5mm}
\label{fig:rq2_params}
\end{figure}

\subsection{RQ2: Parameter Sensitivity of Questions}

\textit{Assuming a fixed STL structure $\varphi$ for a given task type $\mathcal{T}$, do the associated parameters $\boldsymbol{\theta}$ generalize across questions, or are they question-specific?}

Given a shared STL structure $\varphi$ identified for a task type, we investigate whether its continuous parameters $\boldsymbol{\theta}$ can be shared across questions or require instance-level adaptation.
We conceptually compare task-level parameter optimization with question-specific parameter fitting.
If optimal parameters cluster tightly across questions, a single task-level parameterization would suffice; conversely, substantial variation would indicate the need for question-adaptive parameters.

Under a fixed STL structure, we observe pronounced variability in the optimized parameters across individual questions.
As shown in Fig.~\ref{fig:rq2_params}, predicate thresholds and difference-related parameters vary significantly across instances, whereas temporal parameters remain comparatively stable.
This effect is dataset-dependent: BBH exhibits substantially higher parameter dispersion than CLadder, particularly for threshold-related parameters.
Further analysis indicates that this sensitivity is not explained by lexical or semantic similarity between questions.
These results suggest that a single global parameterization is insufficient even when the underlying temporal structure is shared, motivating instance-adaptive parameterization mechanisms that preserve structural interpretability while allowing question-level flexibility, which we explore in Section~\ref{sec:approach}.

\begin{figure}[t]
  \centering
  \includegraphics[width=0.7\columnwidth]{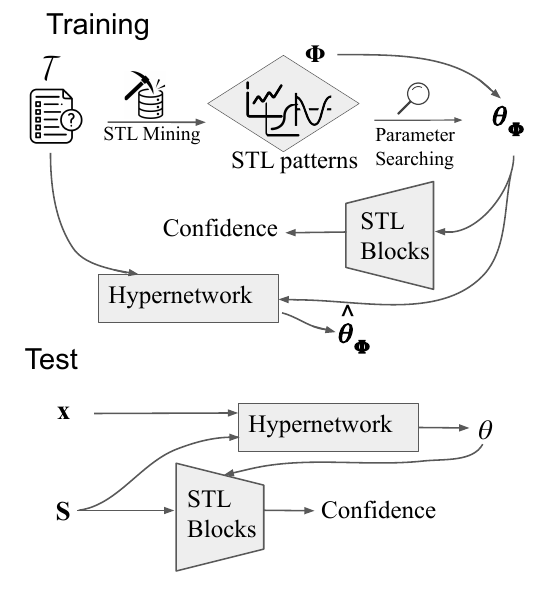}
  \caption{Overview of the confidence quantification pipeline using STL blocks.
During training, discriminative STL mining identifies temporal formula structures, which are instantiated as STL blocks with learnable parameters.
At test time, the learned STL structures are fixed and applied to stepwise confidence signals to produce a scalar confidence score.}
\vspace{-5mm}
  \label{fig:approach}
\end{figure}

\section{Confidence Estimation with STL Blocks}
\label{sec:approach}

% \subsection{STL Blocks Design}
% \label{sec:stl_blocks}

% Given a stepwise confidence signal $\mathbf{S}=[s_1,\dots,s_T]$,
% an STL block evaluates a fixed formula $\varphi$ under standard STL
% robust semantics and produces a scalar \emph{confidence estimate}
% $\hat{p}\in[0,1]$.
% Concretely, the block first computes the aggregated robustness score
% $\rho(\varphi,\mathbf{S})\in\mathbb{R}$, where positive values indicate
% satisfaction of the temporal pattern and negative values indicate violation.
% Since robustness is not probabilistic, we apply a monotonic,
% differentiable mapping to obtain a calibrated confidence score:
% $
% \label{eq:rho2conf_simple}
% \hat{p} = \omega\!\left(\alpha\,\rho(\varphi,\mathbf{S}) + \beta\right),
% $
% where $\omega(\cdot)$ denotes the sigmoid function and
% $\alpha,\beta$ are learnable parameters.
% This mapping preserves the ordering induced by robustness while ensuring
% a bounded, differentiable output suitable for probabilistic training.

\subsection{STL Blocks Design}
\label{sec:stl_blocks}

Given a stepwise confidence signal $\mathbf{S}=[s_1,\dots,s_T]$, an STL block evaluates a fixed STL formula $\varphi$ under quantitative robustness semantics to produce a scalar confidence estimate $\hat{p}\in[0,1]$.
Specifically, the block computes an aggregated robustness score $\rho(\varphi,\mathbf{S})\in\mathbb{R}$ and maps it to a probabilistic confidence via a monotonic, differentiable transformation:
$\hat{p} = \omega\!\left(\alpha\,\rho(\varphi,\mathbf{S}) + \beta\right)$,
where $\omega(\cdot)$ denotes the sigmoid function and $\alpha,\beta$ are learnable parameters.
This mapping preserves robustness ordering while producing a bounded output suitable for probabilistic training.

% This mapping preserves the ordering induced by robustness while ensuring a bounded, differentiable output suitable for probabilistic training.
%The scale parameter $\alpha$ controls sensitivity to robustness variations, and the bias $\beta$ shifts the decision boundary.

\noindent
\textbf{Aggregation across patterns.}
Our design explicitly separates \emph{interpretable temporal structure}, encoded by fixed STL formulas, from \emph{continuous confidence calibration}, controlled by a small set of learnable parameters.
When multiple STL formulas are used, each STL block produces an individual confidence score.
Scores from \emph{positive} patterns (higher indicates correctness) and \emph{negative} patterns (higher indicates incorrectness) are aggregated separately, with negative-pattern scores converted via $1-\hat{p}$ so that all components consistently indicate correctness.
The final confidence estimate is obtained by a weighted aggregation of these components into a single scalar prediction.
Implementation details and numerical stabilizations are deferred to the appendix.

As illustrated in Fig.~\ref{fig:approach}, our confidence estimation pipeline instantiates mined STL formulas as differentiable blocks and applies them to stepwise confidence signals to produce scalar confidence scores.
Our primary approach predicts \emph{question-adaptive parameters} for these STL blocks using a hypernetwork conditioned on the input question.
To assess the necessity of instance-level adaptation, we also consider an ablated variant in which both the STL structure and its parameters are fixed after mining and shared across all test inputs.

\subsection{Question-Adaptive Hypernetworks}
\label{sec:hypernetwork}

A central challenge in question-level confidence estimation is balancing
adaptability with interpretability.
While different questions may require different parameter values, adapting
the temporal structure itself would undermine the interpretability of the
learned STL formulas.
We therefore fix the STL structure discovered during mining and allow only its
continuous parameters to vary across instances.

Given a fixed STL formula structure $\varphi$, we adopt a hypernetwork-based
parameterization to instantiate the parameters of each STL block
on a per-question basis.
This enables question-specific adaptation to varying reasoning
complexity and confidence dynamics while preserving the interpretable temporal
structure encoded by the STL formulas.

\paragraph{Hypernetwork Architecture.}
We fix the STL formula structure $\varphi$ discovered during the mining
stage and use a hypernetwork $\mathcal{H}_{\psi}$ to predict its parameters
on a per-instance basis.
The hypernetwork takes as input the original prompt $\mathbf{x}$ and the
corresponding stepwise confidence signal $\mathbf{S}$, and outputs a
parameter vector:
$
    \theta = \mathcal{H}_{\psi}(\mathbf{x}, \mathbf{S}),
$
where $\theta$ includes all continuous parameters associated with the STL
block, including temporal bounds, predicate thresholds, and the parameters
$\alpha$ and $\beta$ of the robustness-to-confidence mapping.
In practice, the input prompt $\mathbf{x}$ is encoded using a fixed
text encoder, while the confidence signal $\mathbf{S}$ is summarized
using simple temporal statistics (e.g., mean, variance, and slope) or
a lightweight sequence encoder.
The resulting representations are concatenated and fed into a
multi-layer perceptron to produce $\theta_i$.

\paragraph{Training Objective.}
The hypernetwork is trained end-to-end using the same discriminative
objective as in the mining stage.
For each question, the predicted parameters $\theta_i$ are instantiated
into the fixed STL structure $\varphi$, and the resulting robustness
score $\rho(\varphi_{\theta_i}, \mathbf{S}_i)$ is mapped to a confidence
estimate ${P}_i$.
The hypernetwork parameters $\psi$ are learned by minimizing
the discriminative loss over the training set:
$
\min_{\psi} \;\; \frac{1}{N} \sum_{i=1}^{N}
\mathcal{L}\big(c_i, \hat{P}_i\big),
$
where $\hat{P}_i$ is obtained by instantiating the predicted parameters
$\theta_i = \mathcal{H}_{\psi}(\mathbf{x}_i, \mathbf{S}_i)$ into the fixed
STL structure $\varphi$, evaluating the resulting robustness, and applying
the robustness-to-confidence mapping.
The loss $\mathcal{L}$ corresponds to the discriminative objective.

\begin{table*}[t]
\centering
\caption{
Ablation results on uncertainty calibration across base models and datasets.
We report ECE ($\downarrow$) and Brier Score ($\downarrow$) as mean $\pm$ std over 5 folds; best results are in \textbf{bold}.
\textbf{A1}: cross-task reuse of both positive and negative STL patterns (mined on BBH);
\textbf{A2}: cross-task reuse of negative (failure-mode) patterns only;
\textbf{A3}: in-domain STL mining with fixed parameters;
\textbf{Ours}: in-domain STL with instance-adaptive parameters via a hypernetwork.
Since A1 and A2 use the mined pattern from BBH, ``--'' indicates that the A1 and A2 results of BBH are the same with A3.
}
\label{tab:ablation}

\begin{adjustbox}{max width=\textwidth}
\begin{tabular}{@{}ll|cc|cc|cc|cc@{}}
\toprule
\multirow{2}{*}{\textbf{Base Model}} & \multirow{2}{*}{\textbf{Method}}
& \multicolumn{2}{c|}{\textbf{SciQ}}
& \multicolumn{2}{c|}{\textbf{CLadder}}
& \multicolumn{2}{c|}{\textbf{Math}}
& \multicolumn{2}{c}{\textbf{BBH}} \\
\cmidrule(lr){3-4} \cmidrule(lr){5-6} \cmidrule(lr){7-8} \cmidrule(lr){9-10}
& & ECE $\downarrow$ & Brier $\downarrow$
& ECE $\downarrow$ & Brier $\downarrow$
& ECE $\downarrow$ & Brier $\downarrow$
& ECE $\downarrow$ & Brier $\downarrow$ \\
\midrule

% ============================================================================
% Qwen3 Group
% ============================================================================
\multirow{5}{*}{\textbf{Qwen3}}
& AveLogit
& .055$_{\pm.006}$ & .028$_{\pm.004}$
& .200$_{\pm.007}$ & .230$_{\pm.006}$
& .150$_{\pm.023}$ & .177$_{\pm.020}$
& .339$_{\pm.011}$ & .354$_{\pm.009}$\\
& SAR
& .244$_{\pm.039}$ & .272$_{\pm.012}$
& .406$_{\pm.010}$ & .421$_{\pm.009}$
& .373$_{\pm.085}$ & .382$_{\pm.071}$
& .372$_{\pm.011}$ & .382$_{\pm.010}$\\
& Self-Eval
& .010$_{\pm.005}$ & .040$_{\pm.009}$
& .222$_{\pm.005}$ & .250$_{\pm.005}$
& .146$_{\pm.016}$ & .165$_{\pm.018}$
& .197$_{\pm.013}$ & \textbf{.217}$_{\pm.012}$\\
& Self-Consistency
& .046$_{\pm.009}$ & .045$_{\pm.007}$
& .223$_{\pm.016}$ & .239$_{\pm.008}$
& .091$_{\pm.021}$ & \textbf{.143}$_{\pm.020}$
& .248$_{\pm.183}$ & .227$_{\pm.195}$\\
& InternalInspector
& .368$_{\pm.026}$ & .248$_{\pm.057}$
& .236$_{\pm.043}$ & .305$_{\pm.077}$
& \textbf{.064}$_{\pm.016}$ & .168$_{\pm.016}$
& .337$_{\pm.039}$ & .518$_{\pm.050}$ \\
& A1
& .506$_{\pm.005}$ & .283$_{\pm.003}$
& .258$_{\pm.007}$ & .254$_{\pm.001}$
& .354$_{\pm.018}$ & .271$_{\pm.004}$
& -- & -- \\
& A2
& .176$_{\pm.006}$ & .057$_{\pm.004}$
& .084$_{\pm.004}$ & .183$_{\pm.003}$
& .146$_{\pm.015}$ & .168$_{\pm.008}$
& -- & -- \\
& A3
& .005$_{\pm.007}$ & .025$_{\pm.005}$
& .057$_{\pm.006}$ & .178$_{\pm.004}$
& {.082}$_{\pm.009}$ & .158$_{\pm.017}$
& .057$_{\pm.009}$ & .228$_{\pm.003}$ \\
& Ours
& \textbf{.005}$_{\pm.006}$ & \textbf{.025}$_{\pm.005}$
& \textbf{.035}$_{\pm.003}$ & \textbf{.172}$_{\pm.004}$
& {.109}$_{\pm.014}$ & {.154}$_{\pm.014}$
& \textbf{.052}$_{\pm.015}$ & {.224}$_{\pm.003}$ \\

\midrule

% ============================================================================
% Gemma3 Group
% ============================================================================
\multirow{5}{*}{\textbf{Gemma3}}
& AveLogit
& .396$_{\pm.030}$ & .404$_{\pm.023}$
& .127$_{\pm.012}$ & .181$_{\pm.009}$
& .153$_{\pm.028}$ & .171$_{\pm.038}$
& .209$_{\pm.006}$ & .249$_{\pm.004}$\\
& SAR
& .246$_{\pm.039}$ & .265$_{\pm.032}$
& .388$_{\pm.009}$ & .413$_{\pm.009}$
& .381$_{\pm.098}$ & .393$_{\pm.089}$
& .379$_{\pm.011}$ & .400$_{\pm.005}$\\
& Self-Eval
& \textbf{.044}$_{\pm.020}$ & .059$_{\pm.018}$
& .204$_{\pm.011}$ & .222$_{\pm.010}$
& .136$_{\pm.031}$ & \textbf{.135}$_{\pm.043}$
& .220$_{\pm.007}$ & .228$_{\pm.006}$\\
& Self-Consistency
& .052$_{\pm.011}$ & \textbf{.049}$_{\pm.011}$
& .096$_{\pm.024}$ & .228$_{\pm.008}$
& .148$_{\pm.046}$ & .182$_{\pm.036}$
& .186$_{\pm.233}$ & .195$_{\pm.232}$\\
& InternalInspector
& .831$_{\pm.159}$ & .055$_{\pm.064}$
& .172$_{\pm.043}$ & .214$_{\pm.015}$
& \textbf{.060}$_{\pm.040}$ & .167$_{\pm.028}$
& .995$_{\pm.005}$ & \textbf{.001}$_{\pm.001}$\\
& A1
& .142$_{\pm.027}$ & .264$_{\pm.008}$
& .064$_{\pm.008}$ & .171$_{\pm.004}$
& .203$_{\pm.052}$ & .154$_{\pm.018}$
& -- & -- \\
& A2
& .059$_{\pm.022}$ & .246$_{\pm.001}$
& .038$_{\pm.010}$ & .167$_{\pm.006}$
& .157$_{\pm.054}$ & .149$_{\pm.024}$
& -- & -- \\
& A3
& .056$_{\pm.016}$ & .247$_{\pm.001}$
& .042$_{\pm.013}$ & .165$_{\pm.006}$
& {.122}$_{\pm.057}$ & .151$_{\pm.026}$
& .054$_{\pm.013}$ & .196$_{\pm.003}$ \\
& Ours
& {.048}$_{\pm.012}$ & {.245}$_{\pm.003}$
& \textbf{.018}$_{\pm.005}$ & \textbf{.160}$_{\pm.006}$
& {.131}$_{\pm.046}$ & {.141}$_{\pm.024}$
& \textbf{.039}$_{\pm.004}$ & {.189}$_{\pm.004}$ \\

\midrule

% ============================================================================
% Llama Group (placed last)
% ============================================================================
\multirow{5}{*}{\textbf{Llama}}
& AveLogit
& .546$_{\pm.041}$ & .511$_{\pm.031}$
& .402$_{\pm.008}$ & .413$_{\pm.006}$
& .550$_{\pm.070}$ & .530$_{\pm.052}$
& .510$_{\pm.009}$ & .502$_{\pm.006}$\\
& SAR
& .205$_{\pm.016}$ & .258$_{\pm.025}$
& .333$_{\pm.001}$ & .365$_{\pm.005}$
& .292$_{\pm.071}$ & .323$_{\pm.049}$
& .309$_{\pm.012}$ & .330$_{\pm.005}$\\
& Self-Eval
& .056$_{\pm.021}$ & \textbf{.066}$_{\pm.023}$
& .381$_{\pm.011}$ & .383$_{\pm.011}$
& .593$_{\pm.041}$ & .585$_{\pm.042}$
& .454$_{\pm.007}$ & .454$_{\pm.007}$\\
& Self-Consistency
& \textbf{.043}$_{\pm.025}$ & .097$_{\pm.018}$
& .179$_{\pm.025}$ & .273$_{\pm.010}$
& .469$_{\pm.078}$ & .461$_{\pm.059}$
& .242$_{\pm.169}$ & .265$_{\pm.154}$\\
& InternalInspector
& .641$_{\pm.059}$ & .132$_{\pm.039}$
& .026$_{\pm.001}$ & .250$_{\pm.000}$
& .315$_{\pm.046}$ & .231$_{\pm.025}$
& .887$_{\pm.082}$ & \textbf{.039}$_{\pm.008}$\\
& A1
& .064$_{\pm.045}$ & .216$_{\pm.011}$
& .125$_{\pm.009}$ & .266$_{\pm.002}$
& \textbf{.077}$_{\pm.038}$ & .228$_{\pm.020}$
& -- & -- \\
& A2
& .059$_{\pm.020}$ & .214$_{\pm.016}$
& .012$_{\pm.004}$ &  .284$_{\pm.002}$
& .106$_{\pm.045}$ & .229$_{\pm.025}$
& -- & -- \\
& A3
& .058$_{\pm.018}$ & .213$_{\pm.016}$
& .010$_{\pm.004}$ & .285$_{\pm.002}$
& .104$_{\pm.059}$ & .231$_{\pm.027}$
& .014$_{\pm.010}$ & .233$_{\pm.002}$ \\
& Ours
& {.054}$_{\pm.020}$ & {.211}$_{\pm.016}$
& \textbf{.009}$_{\pm.003}$ & \textbf{.233}$_{\pm.001}$
& {.103}$_{\pm.040}$ & \textbf{.225}$_{\pm.023}$
& \textbf{.012}$_{\pm.007}$ & {.232}$_{\pm.002}$ \\

\bottomrule
\end{tabular}
\end{adjustbox}

\vspace{2mm}
% \footnotesize{$^{\dagger}$ ``--'' typically arises when cross-task reuse fails to transfer (random-level performance), yielding degenerate confidence predictions.}
\end{table*}

% ------------------------------------------------------------

\section{Experiments}
\label{sec:experiments}

\begin{figure}[t]
  \includegraphics[width=\columnwidth]{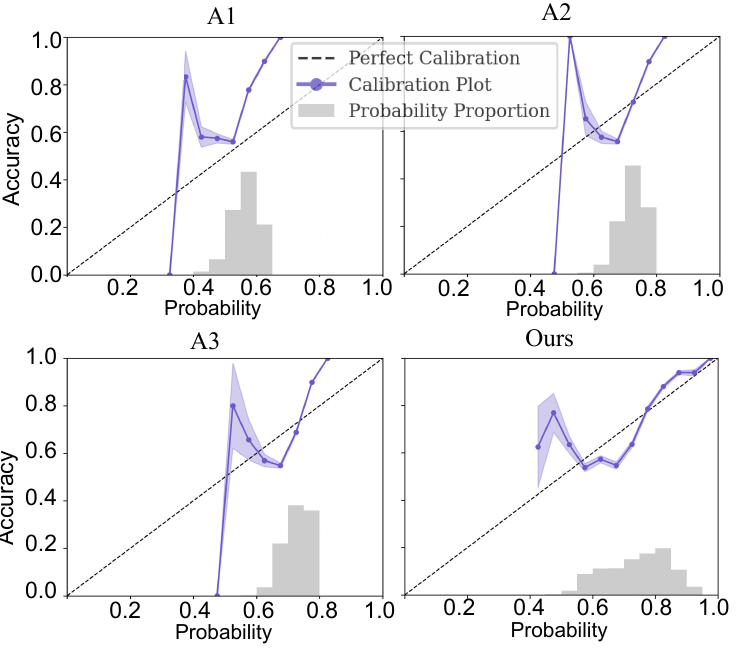}
  \caption{Reliability diagrams for three ablated variants (A1--A3) and the proposed method (Ours) on CLadder with Qwen3.
The dashed diagonal denotes perfect calibration.
Compared to ablations, the proposed method aligns more closely with the diagonal across confidence ranges, indicating improved calibration.}
  \label{fig:relia}
\end{figure}

% ------------------------------------------------------------

\subsection{Experimental Settings}

We evaluate our method on four reasoning benchmarks:
GAOKAO-Math, CLadder, SciQ, and Big-Bench-Hard (BBH).
All tasks are formulated as multiple-choice or binary decision problems, enabling consistent correctness labeling.
Experiments are conducted using three large language models:
Qwen3-8B, Gemma-3-12B, and Llama-3-8B.
Unless otherwise specified, all methods operate in a single-sample inference regime; only self-consistency baselines draw multiple samples.

We compare our STL-based confidence estimator against representative uncertainty estimation baselines, including logit-based~\cite{lin2022teaching}, consistency-based~\cite{portillo-wightman-etal-2023-strength}, self-evaluation~\cite{kadavath2022languagemodelsmostlyknow} approaches and internal-base method~\cite{duan-etal-2024-shifting,beigi-etal-2024-internalinspector}.
All baselines are implemented following their original descriptions and evaluated under identical data splits and inference settings.
We evaluate confidence estimation quality using standard calibration metrics, including Expected Calibration Error (ECE), Brier Score and AUROC.
All metrics are computed on held-out test sets and averaged across folds.

\subsection{Results and Analysis}
\label{sec:ablation}

Table~\ref{tab:ablation} summarizes the main experimental results, analyzing both
cross-task transferability of STL patterns (RQ1) and the effect of
instance-level parameter adaptation (RQ2).

Across backbone models and datasets, reusing only negative (failure-mode) STL
patterns (A2) consistently outperforms full cross-task reuse (A1), confirming
that incorrect-reasoning confidence dynamics transfer more reliably than
correct-reasoning patterns.
However, both cross-task variants are clearly inferior to in-domain STL mining
(A3 and Ours), indicating that task-specific temporal structures and parameters
remain crucial for accurate calibration.
Comparing A3 with Ours isolates the impact of question-level parameter adaptation.
While in-domain fixed STL blocks already yield strong performance, the proposed
hypernetwork-based parameterization further improves calibration in most
settings, with particularly consistent gains on CLadder and BBH.
This demonstrates that, even under a shared temporal structure, confidence
dynamics vary substantially across questions, supporting RQ2.

Standard logit-based, consistency-based, and self-evaluation baselines exhibit
higher calibration error across all benchmarks.
In contrast, STL-based methods (A3 and Ours) achieve more reliable calibration
by explicitly modeling temporal confidence evolution, with Ours providing the
most stable performance under single-sample inference.
Table~\ref{tab:ablation_stl_count} shows that confidence estimation is sensitive
to the number of STL templates.
Too few templates underfit temporal patterns, while too many introduce
redundancy and degrade calibration.
An intermediate configuration (10 balanced templates) achieves the best overall
trade-off across ECE, Brier score, and AUROC.

Despite incorporating logic evaluation and hypernetworks, our method remains
efficient, with an average inference time of $0.55_{\pm 0.04}$ seconds per
example.
This is competitive with single-sample baselines and substantially faster than
sampling-based approaches such as self-consistency.
Overall, the results show that (i) failure-mode STL patterns generalize better
across tasks than success patterns, and (ii) in-domain STL mining combined with
instance-adaptive parameterization yields the most accurate and practical
confidence estimates.

\begin{table}[t]
\centering
\tiny
\caption{Hyperparameter study on the number of STL templates. All configurations use balanced positive and negative templates.}
\label{tab:ablation_stl_count}
\begin{tabular}{lccc}
\toprule
\textbf{STL Configuration} 
& \textbf{ECE}$\downarrow$ 
& \textbf{Brier}$\downarrow$ 
& \textbf{AUROC}$\uparrow$ \\
\midrule
2 templates (1+1, total=2) 
& 0.024$_{\pm .009}$ 
& 0.235$_{\pm .001}$ 
& 0.610$_{\pm .007}$ \\

10 templates (5+5, total=10) 
& \textbf{0.020}$_{\pm .005}$ 
& \textbf{0.234}$_{\pm .001}$ 
& \textbf{0.617}$_{\pm .006}$ \\

16 templates (8+8, total=16) 
& 0.023$_{\pm .007}$ 
& 0.234$_{\pm .001}$ 
& 0.614$_{\pm .008}$ \\
\bottomrule
\end{tabular}
\end{table}

 % \begin{table}[t]
 %  \centering
 %  \caption{Inference time comparison.}
 %  \begin{tabular}{lc}
 %  \toprule
 %  \textbf{Method} & \textbf{Inference Time (ms)} \\
 %  \midrule
 %  AveLogit & -- \\
 %  Self-Eval& -- \\
 %  Self-Consistency& {454,870}$_{\pm267,200}$\\
 %  SAR& -- \\
 %  InternalInspector& \textbf{0.078}$_{\pm0.002}$\\
 %  Baseline 6 & -- \\
 %  \midrule
 %  \textbf{Ours} & {0.55}$_{\pm0.04}$ \\
 %  \bottomrule
 %  \end{tabular}
 %  \end{table}

\section{Conclusion}
\label{sec:conclusion}

We proposed a temporal confidence estimation framework that models
stepwise LLM confidence as a signal evaluated with Signal Temporal Logic.
The resulting confidence scores are both well calibrated and
interpretable.
Experiments show that temporal structures are task-dependent, while
instance-level parameter adaptation via hypernetworks is crucial for
capturing question-specific confidence dynamics.
Overall, our results highlight the role of temporal structure in LLM
confidence estimation and offer an efficient alternative to
sampling-based approaches.

\section{Limitations}
\label{sec:limitations}
Our approach relies on stepwise segmentation of reasoning responses, and the
choice of segmentation granularity may affect the resulting confidence signals.
Moreover, while failure-mode STL patterns transfer across tasks, effective
temporal structures remain largely task-dependent and require in-domain mining.
We also restrict our confidence signals to token-level probabilities for
efficiency and black-box applicability.
Finally, our evaluation focuses on structured reasoning benchmarks, and extending
the framework to more open-ended generation remains future work.

\newpage
\clearpage
\appendix

\noindent
\textbf{Appendix}

% \noindent
% \textbf{Acknowledgement}: We used AI-assisted tools for language polishing and editing of the manuscript. All technical content, experimental design, analysis, and conclusions were developed and verified by the authors.

\section{Relationship Between Confidence and Uncertainty}

In the literature, the terms \emph{uncertainty} and \emph{confidence} are often used interchangeably, but they refer to distinct concepts. Uncertainty (aleatoric or epistemic) typically characterizes ambiguity or lack of knowledge~\cite{kendall2017uncertainties,abdar2021review} in the data distribution or the model. In contrast, confidence refers to a sample-level estimate of the reliability or correctness of a specific generated response. In this work, we focus on confidence estimation rather than explicit modeling of aleatoric or epistemic uncertainty. Uncertainty in deep learning, and specifically in Large Language Models (LLMs), is typically categorized into two primary types: \textit{Aleatoric} and \textit{Epistemic}.
 Often referred to as data uncertainty, \emph{aleatoric uncertainty} arises from the inherent stochasticity within the data distribution itself. In the context of generative models, this typically manifests when an input prompt $\mathbf{x}$ plausibly maps to multiple valid responses $\mathbf{y}$. 
 % For instance, the query \textit{``What is the weather like in New York in winter?''} does not yield a single deterministic answer; rather, factors such as yearly variations and subjective experiences introduce natural ambiguity. 
 Mathematically, this corresponds to a dispersed or multi-modal posterior distribution $p(\mathbf{y}\mid\mathbf{x})$.
 Crucially, aleatoric uncertainty is considered \textit{irreducible}: even with infinite training data and a perfect model, this uncertainty persists as it is an intrinsic property of the underlying task. In this work, we focus on confidence estimation for individual model outputs,
rather than modeling inherent data ambiguity.

Known as model uncertainty, \emph{epistemic uncertainty} stems from the model's lack of knowledge or ignorance regarding the underlying data-generating process. In the context of LLMs, this typically arises when the input prompt falls into ``knowledge blind spots'' or \textit{Out-of-Distribution} (OOD) regions where the training data coverage is sparse or non-existent \citep{malinin2021uncertainty, desai-durrett-2020-calibration}. A paradigmatic example is a query regarding future events, such as \textit{``Where will the 2060 Olympics be held?''}. Since this information is absent from the training corpus, the model is forced to guess, exhibiting high epistemic uncertainty.
Mathematically, this corresponds to uncertainty over the model parameters $\boldsymbol{\theta}$. 
Epistemic uncertainty arises from the model's lack of knowledge about the
data-generating process, for instance, when inputs fall into out-of-distribution
regions or knowledge blind spots. While epistemic uncertainty characterizes the model's lack of knowledge at a distributional level, our work does not aim to explicitly estimate or decompose epistemic uncertainty.
Instead, we focus on confidence estimation at the instance level, i.e., predicting the probability that a specific generated response is correct.
Epistemic uncertainty may influence confidence implicitly, but it is not modeled as a separate uncertainty component.

\subsection{Evaluation Metrics}
To rigorously quantify the quality of the estimated confidence ${P}$, we compare it against the empirical correctness using standard calibration metrics. We primarily utilize the following three metrics:

% \begin{itemize}
\noindent
\textbf{Expected Calibration Error (ECE):} ECE~\cite{guo2017calibration} measures the expected difference between the predicted confidence and the empirical accuracy. It is computed by partitioning samples into $B$ bins based on their confidence scores:
    \begin{equation}
        \text{ECE} = \sum_{b=1}^{B} \frac{|B_b|}{N} \left| \text{acc}(B_b) - \text{conf}(B_b) \right|,
    \end{equation}
    where $N$ is the total number of samples, $|B_b|$ is the number of samples in bin $b$, and $\text{acc}(B_b)$ and $\text{conf}(B_b)$ represent the average accuracy and average confidence within that bin, respectively.

\noindent
\textbf{Brier Score (BS):} The Brier Score~\cite{glenn1950verification} is a strictly proper scoring rule that measures the mean squared error between the predicted probability ${P}_i$ and the binary ground truth outcome $c_i \in \{0, 1\}$ (where 1 indicates a correct response):
    \begin{equation}
        \text{BS} = \frac{1}{N} \sum_{i=1}^{N} ({P}_i - c_i)^2.
    \end{equation}

\section{Signal Temporal Logic Preliminaries}

\subsection{STL Syntax and Semantics} 
Signal Temporal Logic (STL) provides a rigorous formalism for specifying and monitoring the temporal behavior of real-valued signals~\cite{maler2004monitoring}. 
We use STL to describe logical constraints on the evolution of the stepwise confidence signal $\mathbf{S} = [s_1, \dots, s_n]$.

An STL formula is defined by a fixed logical structure $\varphi$ together with a set of continuous parameters $\boldsymbol{\theta}$,
such as predicate thresholds and temporal bounds.
We denote a parameterized STL formula as $\varphi_{\boldsymbol{\theta}}$.
 The basic building block is a \textit{predicate} $\mu$, typically taking the form of an inequality $\mu \equiv s_t \geq c$, where $c$ is a constant threshold. The inequality direction is chosen such that larger (better) values of $s_t$
correspond to higher confidence.%, which aligns with the semantics of the confidence signal considered in this work.

Complex formulas are constructed using Boolean operators (negation $\neg$, conjunction $\wedge$) and temporal operators (Always $\Box$, Eventually $\Diamond$). Our chosen fragment of the STL syntax is given by:
\begin{equation*}
    \varphi ::= \mu \mid \neg \varphi \mid \varphi_1 \land \varphi_2 \mid  \varphi_1 \lor \varphi_2 \mid\Box_{[a,b]} \varphi \mid \Diamond_{[a,b]} \varphi,
\end{equation*}
where $[a, b]$ denotes a time interval with parametric start time $a$ and end $b$. Intuitively, $\Box_{[a,b]} \varphi$ requires $\varphi$ to hold at \textit{all} steps within the interval, while $\Diamond_{[a,b]} \varphi$ requires $\varphi$ to hold at \textit{some} step within the interval. We focus on the $\Box$ and $\Diamond$ operators in this work, as they are
sufficient to capture the temporal confidence patterns of interest,
and omit the Until operator for simplicity.

\noindent
\textbf{STL Robustness.} 
Unlike standard Boolean logic, which yields binary outcomes (True/False), STL offers \textit{quantitative semantics}, also known as robustness degree $\rho$. 
The robustness $\rho(\varphi, \mathbf{S}, t)$ measures \emph{how strongly} the signal $\mathbf{S}$ satisfies the formula $\varphi$ at time $t$. A positive robustness indicates satisfaction, while a negative value implies violation.
For a predicate $\mu \equiv s_t \geq c$, the robustness is simply $\rho(\mu, \mathbf{S}, t) = s_t - c$. 
Crucially, the robustness for temporal operators is computed using min-max operations:
\begin{align}
    \rho(\Box_{[a,b]} \varphi, \mathbf{S}, t) &= \min_{k \in [t+a, t+b]} \rho(\varphi, \mathbf{S}, k), \\
    \rho(\Diamond_{[a,b]} \varphi, \mathbf{S}, t) &= \max_{k \in [t+a, t+b]} \rho(\varphi, \mathbf{S}, k).
\end{align}
This quantitative property is particularly valuable for LLMs, as it transforms logical constraints into differentiable scalar values that can serve as uncertainty metrics or training objectives. As illustrated in Figure~\ref{fig:stl}, STL robustness offers a quantitative way to summarize whether a stepwise confidence trajectory matches a specified temporal pattern.

\section{Primitive STL Templates and Atomic Predicates}
\label{app:templates}

This section summarizes the complete set of primitive STL templates
$\mathcal{T}_{\text{base}}$ and atomic predicates $\mathcal{P}$ used in the
discriminative STL mining procedure described in Section~\ref{sec:rq}.
These templates define the base hypothesis space from which more complex
temporal formulas are constructed through Boolean composition, temporal
nesting, and parameterization.

We emphasize that these primitives are \emph{not} used as fixed rules.
Instead, they serve as a compact, interpretable basis for defining a large
search space of candidate temporal confidence patterns, from which optimal
formulas are selected via data-driven discriminative mining.

\subsection{Primitive STL Templates}
\label{app:primitive_templates}

We group the primitive templates into two categories:
\emph{positive templates} $\varphi^+$, which are intended to capture
temporal confidence patterns characteristic of correct reasoning, and
\emph{negative templates} $\varphi^-$, which describe failure modes such as
instability, confidence collapse, or oscillation.

\paragraph{Positive Templates ($\varphi^+$).}
These templates describe temporal confidence behaviors commonly observed
in correct or well-structured reasoning trajectories.
\begin{table}[h]
\centering
\small
\begin{tabular}{l l}
\toprule
Template & STL Formula \\
\midrule
WeakestLink &
$\Box_{[0,T]}(s_t \geq \mu)$ \\

EndHigh &
$\Box_{[T-k,T]}(s_t \geq \mu)$ \\

StartHigh &
$\Box_{[0,k]}(s_t \geq \mu)$ \\

NeverSharpDrop &
$\Box_{[0,T]}(\Delta_t > -\varepsilon)$ \\

ConfidenceGain &
$\bar{s}_{\text{end}} > \bar{s}_{\text{start}} + \varepsilon$ \\

LowVarOverall &
$\mathrm{Var}(s) \leq \nu$ \\
\bottomrule
\end{tabular}
\caption{Primitive positive STL templates used as base structures for discriminative mining.}
\end{table}

\paragraph{Negative Templates ($\varphi^-$).}
Negative templates are designed to capture temporal confidence patterns
associated with incorrect reasoning, hallucination, or unstable inference.
\begin{table}[h]
\centering
\small
\begin{tabular}{l l}
\toprule
Template & STL Formula \\
\midrule
EventuallyLow &
$\Diamond_{[0,T]}(s_t \leq \mu)$ \\

EndLow &
$\Diamond_{[T-k,T]}(s_t \leq \mu)$ \\

ConfidenceLoss &
$\bar{s}_{\text{start}} > \bar{s}_{\text{end}} + \varepsilon$ \\

SharpDrop &
$\Diamond_{[0,T]}(\Delta_t \leq -\varepsilon)$ \\

FinalDecline &
$\Box_{[T-k,T]}(\Delta_t \leq 0)$ \\

Recovery &
$\Diamond(s_t \leq \mu_l) \land \Diamond(s_t \geq \mu_h)$ \\
\bottomrule
\end{tabular}
\caption{Primitive negative STL templates capturing failure-related confidence patterns.}
\end{table}

For each template, we optimize its parameters $\theta$
(in the above examples, $\theta = (a,b,\mu)$) by minimizing a classification-based objective
that separates correct and incorrect responses.
Specifically, we compute the robustness score
$\rho(\varphi,\mathbf{S})$ for each response and 
 minimize negative log-likelihood (NLL) loss with respect to the correctness labels.
This objective is identical in form to the loss used later for confidence estimation,
but is applied here solely for evaluating the discriminative utility of candidate STL templates.
This step yields a set of \textit{parameterized base templates}
$\{(\varphi_k, \theta_k)\}$, each instantiated from a predefined template
and fitted to the data to maximize its discriminative power.

\noindent
\textbf{Robustness-Based Signal Augmentation.}
To enable the construction of nested STL formulas without explicitly
enumerating complex parse trees, we adopt a robustness-based signal
augmentation strategy, commonly referred to as \textit{signal lifting} in the
STL mining literature~\cite{jha2019telex,donze2010robust,kong2014temporal}.
For each parameterized base template $(\varphi_k, \theta_k)$ discovered in
Step~1, we compute the robustness signal
$r_j \in \mathbb{R}^n$ defined by $r_j[t] = \rho(\varphi_j, \mathbf{S}, t)$
for all time steps $t$.

We then construct an augmented multi-dimensional signal
$\mathbf{S}' \in \mathbb{R}^{n \times (d + k)}$ by concatenating these
robustness signals along the feature dimension:
\begin{equation}
    \mathbf{S}' = \left[ \mathbf{S} \;\middle|\; r_1 \;\middle|\; r_2 \;\middle|\; \dots \;\middle|\; r_k \right].
\end{equation}

Based on this augmented signal, we define new atomic predicates indicating the satisfaction of the inner formula: $\mu_j^+ := (r_j \geq 0)$ (formula satisfied) and $\mu_j^- := (r_j < 0)$ (formula violated).

\noindent
\textbf{Structural Composition.}
Using the augmented predicates $\mu_j^+$ and $\mu_j^-$, we generate complex structures via two mechanisms:
\begin{itemize}
\looseness=-1
    \item \textbf{Temporal Nesting:} We apply temporal templates to the new predicates (e.g., $\Diamond_{[a,b]} \mu_j^+$), which is equivalent to applying the same
temporal operator to the original formula $\phi_j$, i.e.,
$\Diamond_{[a,b]} \phi_j$, under the robustness semantics.
    \item \textbf{Boolean Combination:} To cover diverse error patterns, we combine formulas using logic operators. Negation is implicitly handled via the complementary predicates
$\mu_j^+$ and $\mu_j^-$, corresponding to satisfaction and violation
of the underlying formula, respectively.
Specifically, if a set of formulas $\{\phi_k\}$ effectively captures distinct subsets of errors, we combine them via disjunction $\bigvee \phi_k$. Conversely, to tighten the criteria for correctness, we employ conjunction $\bigwedge \phi_k$.
\end{itemize}

\noindent
\textbf{Dual-Class Template Mining.}
Using the candidate STL formulas generated through Steps~1--3,
we explicitly perform dual-class discriminative mining to identify
two complementary categories of temporal patterns.
Here, $z_i(\phi)$ is obtained by applying a simple monotonic transformation to the scalar robustness
$\rho(\phi, \mathbf{S}_i)$, so that larger robustness yields a larger logit.
Building on the discriminative objective used for parameterizing base templates in Step~1, we explicitly search for two complementary categories of patterns.
We aim to find formulas that not only recognize their target class but also suppress the non-target class:

\noindent
\textit{Correct Patterns ($\Phi_{\text{pos}}$):}
    These formulas characterize correct reasoning flows (e.g., ``confidence
    consistently increases'').
    They are optimized to assign high confidence to correct responses and low
    confidence to the incorrect ones. 
In this setting, $z_i(\phi)$ denotes a discriminative logit for the target class of $\phi$:
larger values indicate stronger evidence that instance $i$ belongs to the target class.

    Using the empirical correctness label $c_i \in \{0,1\}$, we minimize the
    standard Negative Log-Likelihood (NLL):
\begin{equation}
\tiny
\mathcal{L}_{\text{pos}}
= - \frac{1}{N} \sum_{i=1}^{N}
\Big[
c_i \log \sigma\!\big(z_i(\phi)\big)
+ (1 - c_i) \log \big(1 - \sigma\!\big(z_i(\phi)\big)\big)
\Big],
\label{eq:Lpos}
\end{equation}

\noindent
\textit{Incorrect Patterns ($\Phi_{\text{neg}}$):}
    These formulas capture hallucinations or structural errors
    (e.g., ``confidence oscillates'').
    For such patterns, a high robustness score should indicate a high
    probability of incorrectness.
    For $\Phi_{\text{neg}}$, the target class is incorrectness; equivalently, we flip the label and reuse the same logistic mapping on $z_i(\phi)$.
    Accordingly, we invert the target label and optimize:

\begin{equation}
\tiny
\mathcal{L}_{\text{neg}}
= - \frac{1}{N} \sum_{i=1}^{N}
\Big[
(1 - c_i) \log \sigma\!\big(z_i(\phi)\big)
+ c_i \log \big(1 - \sigma\!\big(z_i(\phi)\big)\big)
\Big],
\label{eq:Lneg}
\end{equation}

We hypothesize that correct and incorrect reasoning traces exhibit distinct temporal confidence signatures.
We capture such signatures using STL formulas evaluated under quantitative robustness semantics,
using both a scalar robustness score $\rho(\varphi, \mathbf{S}_i)$ for discriminative scoring and
a time-resolved robustness signal $\rho(\varphi, \mathbf{S}_i, t)$ for signal lifting in Step~2.
Following prior STL mining work~\citep{jha2019telex}, we adapt these techniques to the LLM confidence setting
and learn discriminative STL patterns that separate correct from incorrect responses.

Formally, given a dataset $\{(\mathbf{S}_i, c_i)\}_{i=1}^{N}$, we mine STL formulas $\phi$
whose robustness separates the two classes.
For a candidate formula $\phi$, we compute a robustness-derived logit $z_i(\phi)$ and optimize
a NLL objective on $\sigma(z_i(\phi))$ to quantify its discriminative utility.
This can be expressed as the following optimization problem:
\begin{equation}
(\varphi^*, \theta^*) = 
\arg\min_{\varphi, \theta}
\; \frac{1}{N} \sum_{i=1}^N 
\mathcal{L}\big(c_i, \hat{P}_i\big),
\label{eq:stl_opt}
\end{equation}
where $\hat{P}_i$ denotes the predicted confidence for instance $i$ and $\mathcal{L}$ corresponds to the discriminative loss defined in
Eq.~(12)–(13).

This formulation allows the temporal logic structure $\varphi$ and its parameters $\theta$ to be learned under a calibration-oriented objective, while treating STL robustness as a structured confidence signal rather than a probability itself.

Having performed STL mining, we study the generalizability of the mined patterns from two complementary analytical perspectives.
 Figure~\ref{fig:rqs} provides a schematic overview of the two research questions, illustrating the distinction between structural generalization across task types (RQ1) and parameter sensitivity under a fixed formula (RQ2).

\subsection{Atomic Predicates}
\label{app:atomic_predicates}

The primitive templates are instantiated over a shared set of atomic
predicates $\mathcal{P}$, defined over the stepwise confidence signal
$\mathbf{S} = [s_1, \dots, s_n]$ and its first-order difference
$\Delta_t = s_t - s_{t-1}$.

\begin{table}[h]
\centering
\small
\begin{tabular}{l l}
\toprule
Category & Predicate \\
\midrule
Level &
$s_t \geq \mu$, \quad $s_t \leq \mu$ \\

Deviation &
$s_t \geq \bar{s} - k\sigma$ \\

Derivative &
$\Delta_t > -\varepsilon$, \quad
$\Delta_t \leq -\varepsilon$, \quad
$\Delta_t \leq 0$ \\

Aggregate &
$\mathrm{Var}(s) \leq \nu$, \quad
$\bar{s}_{\text{end}} > \bar{s}_{\text{start}}$ \\
\bottomrule
\end{tabular}
\caption{Atomic predicates used to instantiate STL templates.}
\end{table}

\subsection{Compositional Search Space}

Although the base template set $\mathcal{T}_{\text{base}}$ contains only
14 primitive templates, it induces a large hypothesis space through
systematic composition:
(i) Boolean combination (conjunction and disjunction),
(ii) temporal nesting of operators, and
(iii) parameterized time windows and thresholds.

For example, pairwise Boolean composition alone yields
$\binom{14}{2} = 91$ conjunctions and 91 disjunctions, and higher-order
compositions further expand the space.
When combined with multiple temporal windows and continuous parameter
ranges, the effective search space quickly reaches hundreds or thousands
of candidate formulas.

This combinatorial growth motivates the use of discriminative STL mining
to automatically identify a small set of informative temporal confidence
patterns from data.

\section{Implementation Details}
\label{app:implementation}

\subsection{STL Blocks and Robustness-to-Confidence Mapping}
\label{app:stl_blocks}

This appendix provides implementation-level details for the STL blocks
introduced in Section~\ref{sec:stl_blocks}.
These details are omitted from the main text for clarity and reproducibility.

\paragraph{Robustness computation.}
Each STL block evaluates a fixed STL formula $\varphi$ on a stepwise confidence
signal $\mathbf{S}=[s_1,\dots,s_T]$ and produces both an aggregated robustness
value $\rho(\varphi,\mathbf{S})$ and, optionally, a per-step robustness trace
$\rho(\varphi,\mathbf{S},t)$.
Temporal operators are implemented using standard soft approximations of STL
robust semantics, such as soft-min and soft-max, to ensure differentiability.

\paragraph{Numerical stabilization.}
In practice, robustness values may attain large magnitudes, which can lead to
numerical instability when passed through nonlinear mappings.
To mitigate this, aggregated robustness values are clipped to a bounded range
prior to further processing.
After probability mapping, outputs are additionally clipped to avoid saturation
at $0$ or $1$, ensuring stable optimization under binary cross-entropy loss.

\paragraph{Parameterized sigmoid mapping.}
The robustness-to-confidence transform is implemented as a learnable sigmoid:
\begin{equation}
\hat{p} = \omega(\alpha\,\tilde{\rho} + \beta),
\end{equation}
where $\tilde{\rho}$ denotes the clipped robustness value.
The scale parameter $\alpha$ controls the sharpness of the decision boundary,
while the bias parameter $\beta$ controls its location.
Both parameters are optimized jointly with the rest of the model.

\paragraph{Interpretation of mapping parameters.}
Smaller values of $\alpha$ yield smoother transitions between low and high
confidence, making the estimator less sensitive to small robustness variations.
Larger values of $\alpha$ produce sharper transitions, approaching a hard
threshold on robustness.
The bias parameter $\beta$ determines the robustness level at which the
predicted confidence crosses $0.5$.

\paragraph{Aggregation over multiple STL formulas.}
When multiple STL formulas are active, each produces an individual confidence
score.
Scores from formulas associated with correct reasoning patterns
($\Phi_{\text{pos}}$) are treated as positive evidence, while scores from
incorrect reasoning patterns ($\Phi_{\text{neg}}$) are inverted to represent
negative evidence.
The final confidence estimate is obtained by a weighted aggregation of these
components, as described in Section~\ref{sec:stl_blocks}.

\section{Additional Results}

\subsection{Analysis of Question Similarity and Parameter Variability}

To investigate whether parameter variability can be explained by the intrinsic similarity of questions within a dataset, we analyze the intra-dataset question similarity using both TF-IDF and sentence-transformer-based embeddings.
As reported in Table~\ref{tab:question_similarity}, datasets with higher question similarity do not necessarily exhibit lower parameter variance, and vice versa.
For example, BBH exhibits relatively high semantic similarity between questions, yet shows substantial parameter variability, whereas CLadder displays lower question similarity but more stable parameter configurations.

\subsection{Representative Mined STL Patterns}
\label{app:mined_stl}

This appendix presents representative Signal Temporal Logic (STL) formulas
\emph{discovered by the mining procedure} described in Section~\ref{sec:rq}.
Rather than listing the full search space, we report the most frequently
occurring and consistently generalizable patterns observed across datasets
and backbone models.

These formulas correspond to empirical confidence dynamics that repeatedly
emerge during mining, particularly those associated with strong cross-task
structural reuse (RQ1).
They are reported here to complement the quantitative analyses in the main
text and to improve interpretability.

\begin{table}[htbp]
\centering
\small
\caption{RQ1: Average Jaccard similarity of mined STL templates across task types.
Results are reported as mean $\pm$ standard deviation over folds, separately for correct and incorrect STL templates and for different backbone models.
Incorrect templates consistently exhibit higher cross-task similarity than correct templates, indicating stronger structural reuse.
}
\label{tab:rq1_similarity_gsm8k}
\begin{tabular}{lcc}
\toprule
\textbf{Model} & \textbf{Correct Templates} & \textbf{Incorrect Templates} \\
\midrule
Qwen & $0.473 \pm 0.040$ & $0.811 \pm 0.045$ \\
Gemma3 & $0.549 \pm 0.050$ & $0.789 \pm 0.108$ \\
Llama & $0.532 \pm 0.057$ & $0.744 \pm 0.027$ \\
\bottomrule
\end{tabular}
\end{table}

\begin{table*}[t]
\centering
\small
\caption{
Representative STL formulas discovered by discriminative mining.
Positive patterns characterize confidence dynamics associated with correct
reasoning, while negative patterns capture common failure modes.
Here, $s_t$ denotes the stepwise confidence at reasoning step $t$,
$\Delta_t = s_t - s_{t-1}$ denotes the confidence change between steps,
$T$ is the total number of reasoning steps,
and $\mu,\varepsilon$ are learned thresholds.
}
\label{tab:mined_stl_patterns}
\begin{tabular}{lll}
\toprule
\textbf{Category} & \textbf{Pattern Name} & \textbf{Discovered STL Formula} \\
\midrule
\multirow{4}{*}{Positive}
& WeakestLink
& $\Box_{[0,T]}(s_t \geq \mu)$ \\

& EndHigh
& $\Box_{[T-k,T]}(s_t \geq \mu)$ \\

& StartHigh
& $\Box_{[0,k]}(s_t \geq \mu)$ \\

& NeverSharpDrop
& $\Box_{[0,T]}(\Delta_t > -\varepsilon)$ \\

\midrule
\multirow{4}{*}{Negative}
& EventuallyLow
& $\Diamond_{[0,T]}(s_t \leq \mu)$ \\

& EndLow
& $\Diamond_{[T-k,T]}(s_t \leq \mu)$ \\

& SharpDrop
& $\Diamond_{[0,T]}(\Delta_t \leq -\varepsilon)$ \\

& Recovery
& $\Diamond(\text{low}) \land \Diamond(\text{high})$ \\
\bottomrule
\end{tabular}
\end{table*}

\begin{table*}[t]
\centering
\caption{
Ablation study on uncertainty quantification: AUROC ($\uparrow$) with mean $\pm$ std over 5 folds.
Higher is better. 
}
\label{tab:ablation_auroc}

\begin{tabular}{@{}ll|c|c|c|c@{}}
\toprule
\textbf{Base Model} & \textbf{Variant}
& \textbf{SciQ}
& \textbf{CLadder}
& \textbf{Math}
& \textbf{BBH} \\
\midrule

% ============================================================================
% Qwen3 Group
% ============================================================================
\multirow{5}{*}{\textbf{Qwen3}}
& AveLogit
& .505$_{\pm.068}$
& .718$_{\pm.005}$
& .800$_{\pm.024}$
& .536$_{\pm.006}$\\
& Self-Eval
& .525$_{\pm.072}$
& .534$_{\pm.016}$
& .540$_{\pm.045}$
& .540$_{\pm.017}$\\
& Self-Consistency
& {.789}$_{\pm.042}$
& .726$_{\pm.023}$
& .765$_{\pm.047}$
& .595$_{\pm.210}$\\
& InternalInspector
& .514$_{\pm.058}$
& .576$_{\pm.085}$
& .503$_{\pm.009}$
& .548$_{\pm.002}$\\
& A1
& .502$_{\pm.150}$
& .685$_{\pm.004}$
& .737$_{\pm.035}$
& -- \\
& A2
& .509$_{\pm.171}$
& .713$_{\pm.003}$
& .731$_{\pm.035}$
& -- \\
& A3
& .501$_{\pm.139}$
& .719$_{\pm.013}$
& .749$_{\pm.033}$
& .643$_{\pm.003}$ \\
& Ours
& .521$_{\pm.150}$
& {.735}$_{\pm.007}$
& {.777}$_{\pm.030}$
& {.650}$_{\pm.001}$ \\

\midrule

% ============================================================================
% Gemma3 Group
% ============================================================================
\multirow{4}{*}{\textbf{Gemma3}}
& AveLogit
& .551$_{\pm.024}$
& .644$_{\pm.014}$
& .821$_{\pm.066}$
& .548$_{\pm.014}$\\
& Self-Eval
& {.601}$_{\pm.111}$
& .599$_{\pm.022}$
& {.854}$_{\pm.073}$
& .629$_{\pm.020}$\\
& Self-Consistency
& .586$_{\pm.065}$
& .597$_{\pm.018}$
& .797$_{\pm.138}$
& {.675}$_{\pm.215}$\\
& InternalInspector
& .768$_{\pm.403}$
& .527$_{\pm.034}$
& .501$_{\pm.041}$
& .663$_{\pm.337}$\\
& A1
& .583$_{\pm.029}$
& .612$_{\pm.012}$
& .832$_{\pm.075}$
& -- \\
& A2
& .584$_{\pm.026}$
& .626$_{\pm.015}$
& .825$_{\pm.077}$
& -- \\
& A3
& .580$_{\pm.028}$
& .646$_{\pm.019}$
& .832$_{\pm.078}$
& .657$_{\pm.026}$ \\
& Ours
& .581$_{\pm.032}$
& {.667}$_{\pm.013}$
& .836$_{\pm.077}$
& .661$_{\pm.025}$ \\

\midrule

% ============================================================================
% Llama Group (placed last)
% ============================================================================
\multirow{4}{*}{\textbf{Llama}}
& AveLogit
& .529$_{\pm.045}$
& .507$_{\pm.014}$
& .511$_{\pm.090}$
& .567$_{\pm.015}$\\
& Self-Eval
& .573$_{\pm.106}$
& .526$_{\pm.014}$
& {.585}$_{\pm.111}$
& .523$_{\pm.013}$\\
& Self-Consistency
& {.628}$_{\pm.077}$
& {.553}$_{\pm.032}$
& .554$_{\pm.100}$
& .554$_{\pm.178}$\\
& InternalInspector
& .504$_{\pm.197}$
& .507$_{\pm.016}$
& .527$_{\pm.039}$
& .715$_{\pm.024}$\\
& A1
& .571$_{\pm.060}$
& .504$_{\pm.015}$
& .549$_{\pm.092}$
& -- \\
& A2
& .563$_{\pm.058}$
& .512$_{\pm.012}$
& .547$_{\pm.096}$
& -- \\
& A3
& .554$_{\pm.046}$
& .513$_{\pm.011}$
& .555$_{\pm.094}$
& .581$_{\pm.016}$ \\
& Ours
& .557$_{\pm.035}$
& .519$_{\pm.012}$
& .569$_{\pm.087}$
& {.586}$_{\pm.019}$ \\

\bottomrule
\end{tabular}

\vspace{2mm}
\end{table*}

\begin{figure*}[t]
  \includegraphics[width=2.1\columnwidth]{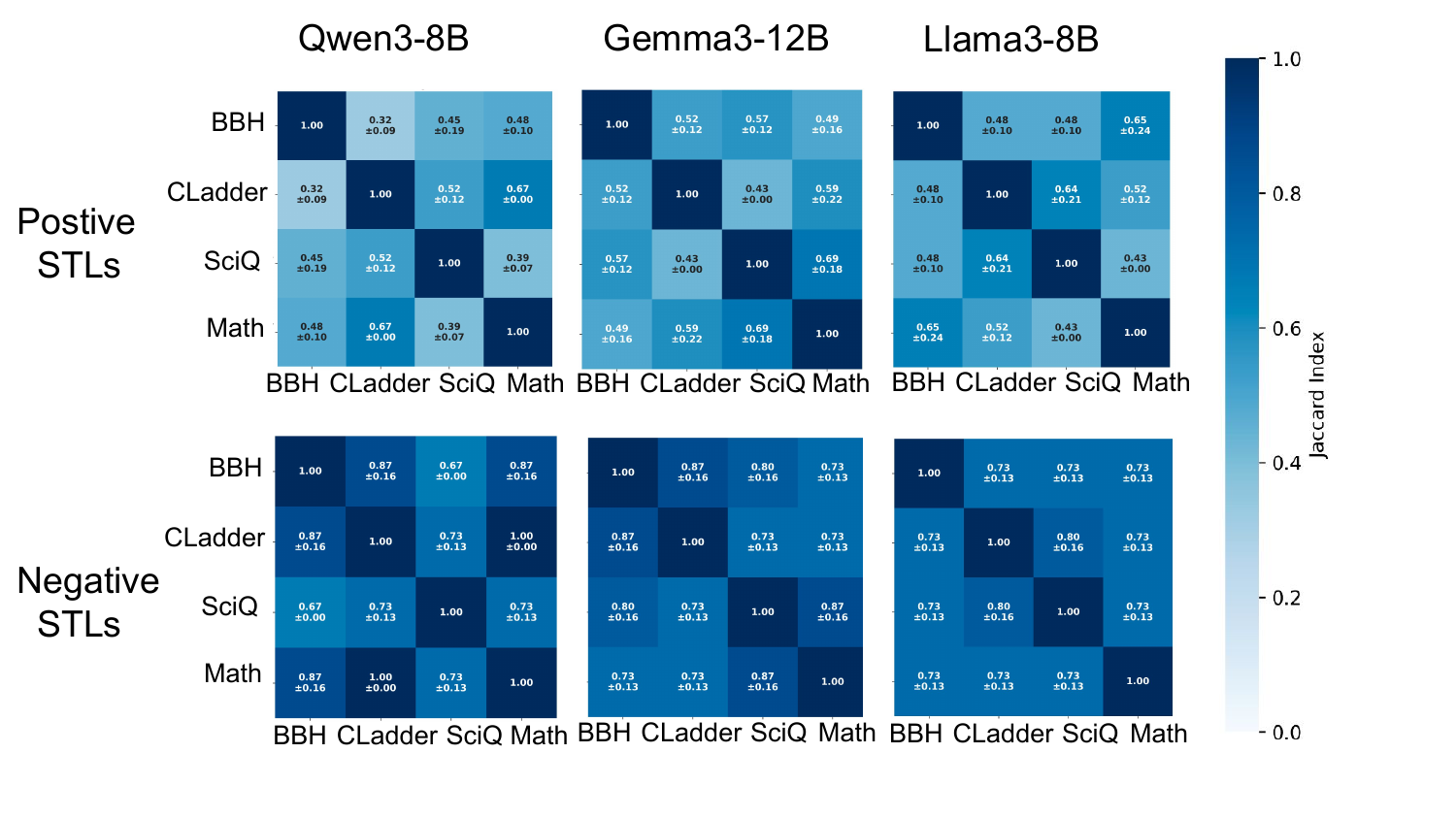}
\caption{
RQ1: Pairwise similarity of mined STL formulas across four reasoning benchmarks for different backbone models.
Results are shown separately for correct (top) and incorrect (bottom) STL patterns, using Qwen3-8B, Gemma3-12B, and Llama3-8B.
Across all models, incorrect STL patterns consistently exhibit higher cross-task similarity than correct patterns, confirming that the asymmetric generalization behavior observed in the main text is not model-specific.
}
\label{fig:rq1_all_models}
\end{figure*}

\begin{figure*}[t]
  \includegraphics[width=2.1\columnwidth]{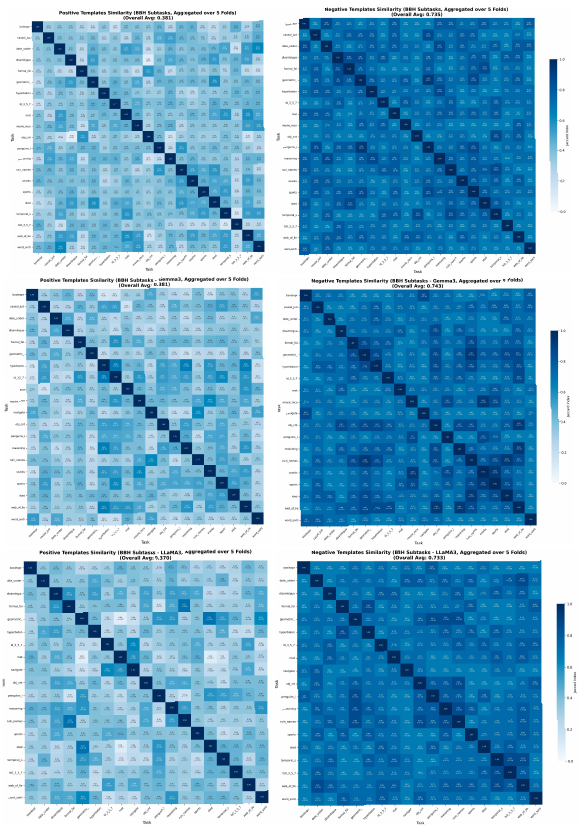}
\caption{
Pairwise similarity of mined STL formulas across BBH subtasks.
Results are aggregated over five folds and shown separately for correct (left) and incorrect (right) STL patterns, across different backbone models.
Correct STL patterns exhibit low similarity across subtasks, while incorrect patterns remain highly consistent.
This trend mirrors the dataset-level analysis and suggests asymmetric generalization behavior at finer task granularity.
}
\label{fig:rq1_bbh}
\end{figure*}

\begin{table*}[htbp]
  \centering
  \tiny
  \caption{Cross-Fold Parameter MAE (Mean $\pm$ Std) for Different Models and Datasets}
  \label{tab:param_mae}
  \resizebox{\textwidth}{!}{%
  \begin{tabular}{llcccc}
  \toprule
  Type & Model & CLadder & BBH & sciq & math \\
  \midrule
  \multirow{3}{*}{Time step} & Qwen3 & $0.204 \pm 0.084$ & $6.534 \pm 4.390$ & $1.070 \pm 0.692$ & $0.221 \pm 0.091$ \\
   & Llama & $3.082 \pm 1.758$ & $0.985 \pm 0.597$ & $1.629 \pm 1.061$ & $0.882 \pm 0.462$ \\
   & Gemma & $1.597 \pm 1.592$ & $2.265 \pm 1.506$ & $1.709 \pm 0.785$ & $1.875 \pm 0.819$ \\
  \midrule
  \multirow{3}{*}{Threshold} & Qwen3 & $0.010 \pm 0.003$ & $0.011 \pm 0.003$ & $0.033 \pm 0.026$ & $0.023 \pm 0.009$ \\
   & Llama & $0.025 \pm 0.011$ & $0.018 \pm 0.008$ & $0.041 \pm 0.023$ & $0.043 \pm 0.025$ \\
   & Gemma & $0.026 \pm 0.014$ & $0.020 \pm 0.011$ & $0.037 \pm 0.016$ & $0.038 \pm 0.020$ \\
  \midrule
  \multirow{3}{*}{Difference} & Qwen3 & $0.010 \pm 0.003$ & $0.004 \pm 0.002$ & $0.010 \pm 0.006$ & $0.017 \pm 0.009$ \\
   & Llama & $0.007 \pm 0.004$ & $0.011 \pm 0.006$ & $0.012 \pm 0.007$ & $0.034 \pm 0.018$ \\
   & Gemma & $0.013 \pm 0.005$ & $0.006 \pm 0.003$ & $0.013 \pm 0.007$ & $0.020 \pm 0.011$ \\
  \bottomrule
  \end{tabular}%
  }
  \end{table*}

\begin{table}[htbp]
  \centering
\caption{
Intra-dataset question similarity measured using TF-IDF and sentence-transformer embeddings.
Values are reported as mean and standard deviation of pairwise cosine similarity within each dataset.
Higher similarity does not necessarily correspond to lower STL parameter variability, suggesting that parameter sensitivity is not solely explained by surface-level or semantic similarity among questions.
}
\label{tab:question_similarity}
  \begin{tabular}{lcccc}
  \toprule
  \multirow{2}{*}{Dataset} & \multicolumn{2}{c}{TF-IDF} & \multicolumn{2}{c}{Sentence Transformers} \\
  \cmidrule(lr){2-3} \cmidrule(lr){4-5}
   & Mean & Std & Mean & Std \\
  \midrule
  CLadder & $0.054$ & $0.125$ & $0.276$ & $0.172$ \\
  BBH & $0.181$ & $0.136$ & $0.767$ & $0.197$ \\
  sciq & $0.565$ & $0.093$ & $0.716$ & $0.062$ \\
  mathnpz & $0.103$ & $0.168$ & $0.405$ & $0.144$ \\
  \bottomrule
  \end{tabular}
  \end{table}

  \begin{figure*}[t]
    \centering

    \begin{subfigure}[t]{0.7\textwidth}
        \centering
        \includegraphics[width=\linewidth]{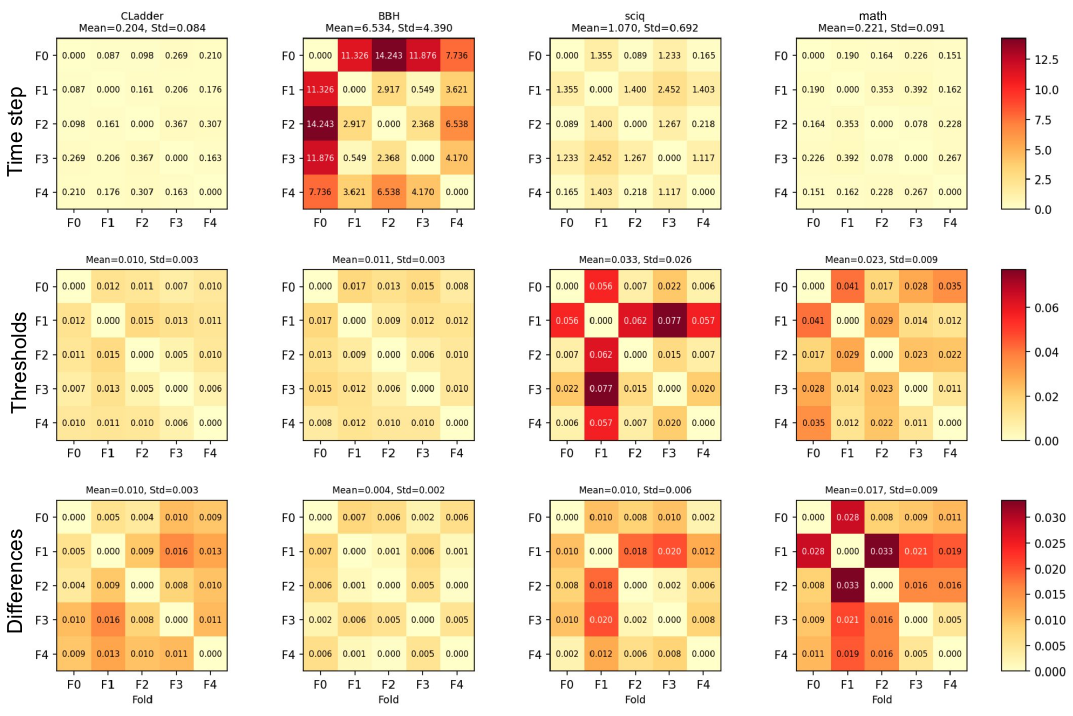}
        \caption{Qwen3}
        \label{fig:params_time}
    \end{subfigure}

    \vspace{0.6em}

    \begin{subfigure}[t]{0.7\textwidth}
        \centering
        \includegraphics[width=\linewidth]{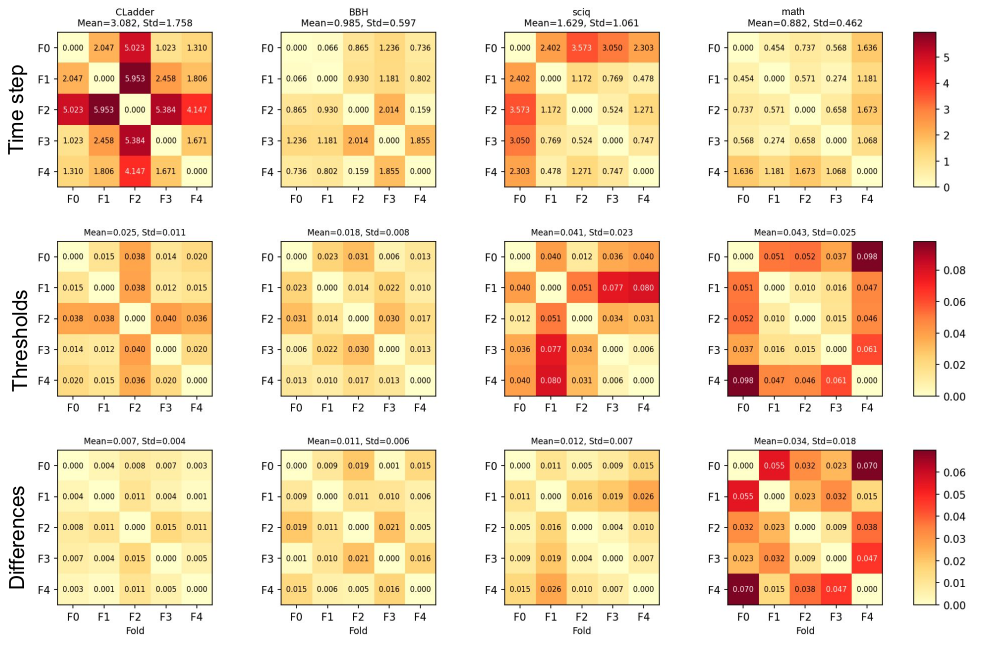}
        \caption{Gemma3}
        \label{fig:params_threshold}
    \end{subfigure}

    \vspace{0.6em}

    \begin{subfigure}[t]{0.7\textwidth}
        \centering
        \includegraphics[width=\linewidth]{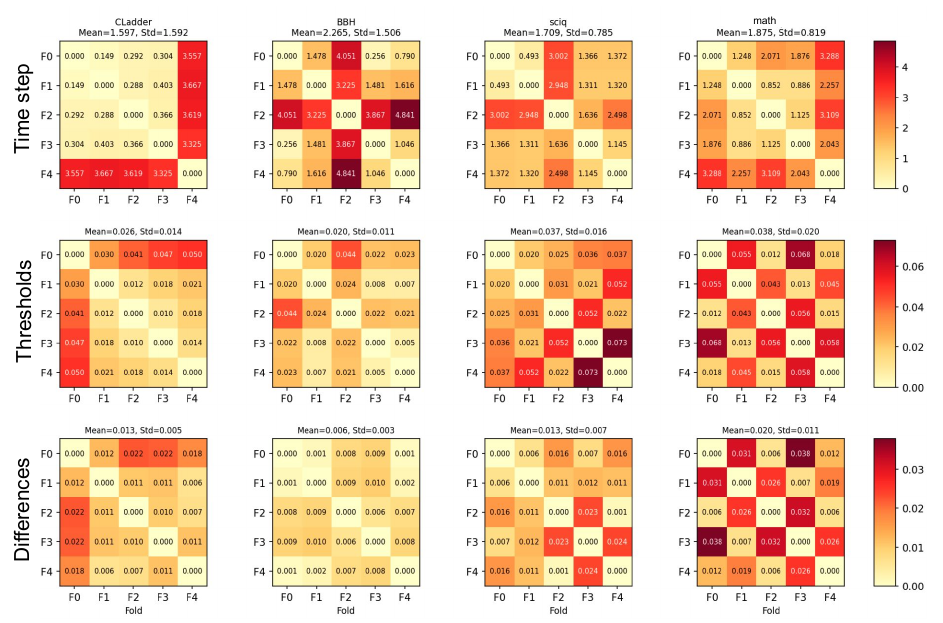}
        \caption{Llama3}
        \label{fig:params_difference}
    \end{subfigure}

    \caption{
    Instance-level STL parameter variability under a fixed temporal structure.
    Each row corresponds to a different parameter category.
    }
    \label{fig:appendix_params_3rows}
\end{figure*}

\begin{figure*}[t]
    \centering

    \begin{minipage}{0.95\textwidth}
        \centering
        \includegraphics[width=\linewidth]{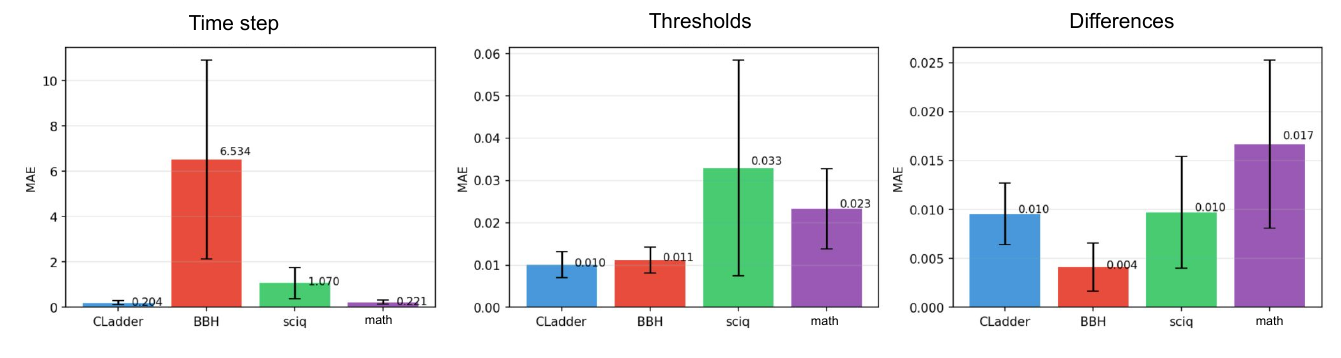}
        \subcaption{Qwen3}
        \label{fig:row_time}
    \end{minipage}

    \vspace{0.6em}

    \begin{minipage}{0.95\textwidth}
        \centering
        \includegraphics[width=\linewidth]{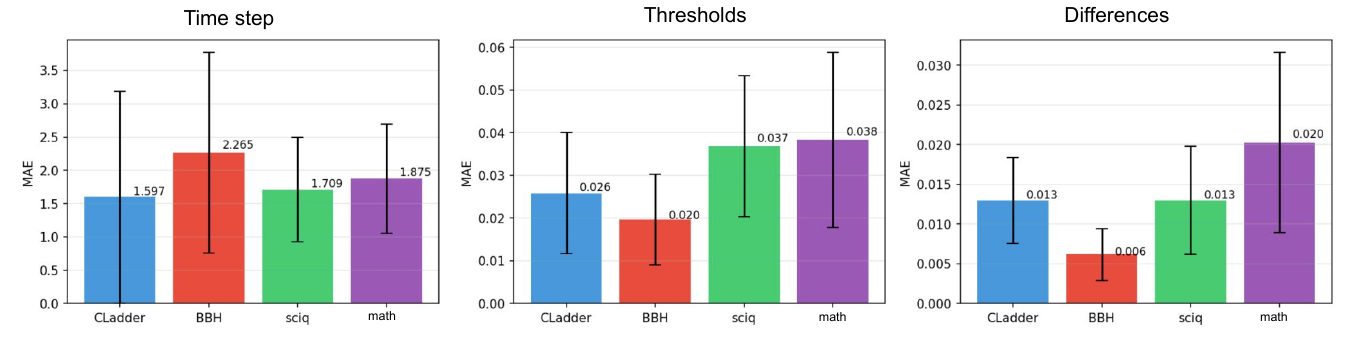}
        \subcaption{Gemma3}
        \label{fig:row_threshold}
    \end{minipage}

    \vspace{0.6em}

    \begin{minipage}{0.95\textwidth}
        \centering
        \includegraphics[width=\linewidth]{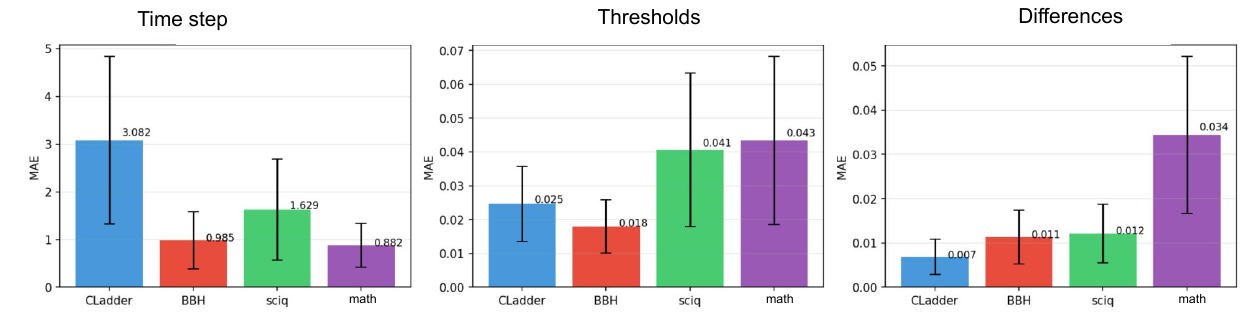}
        \subcaption{Llama3}
        \label{fig:row_difference}
    \end{minipage}
\caption{
RQ2: Question-level sensitivity of optimized STL parameters under a fixed formula structure.
We report the mean absolute difference (MAE) of learned parameters across individual questions, grouped by parameter type: temporal bounds (left), predicate thresholds (middle), and difference-related parameters (right).
Results are shown for four datasets across three backbone models (Qwen3, Gemma3, Llama3).
While temporal parameters remain relatively stable, threshold and difference parameters exhibit substantially higher variability across questions, particularly on BBH and CLadder.
}

    \label{fig:appendix_params_minipage}
\end{figure*}


\begin{thebibliography}{51}
\providecommand{\natexlab}[1]{#1}

\bibitem[{Abdar et~al.(2021)Abdar, Pourpanah, Hussain, Rezazadegan, Liu, Ghavamzadeh, Fieguth, Cao, Khosravi, Acharya et~al.}]{abdar2021review}
Moloud Abdar, Farhad Pourpanah, Sadiq Hussain, Dana Rezazadegan, Li~Liu, Mohammad Ghavamzadeh, Paul Fieguth, Xiaochun Cao, Abbas Khosravi, U~Rajendra Acharya, and 1 others. 2021.
\newblock A review of uncertainty quantification in deep learning: Techniques, applications and challenges.
\newblock \emph{Information fusion}, 76:243--297.

\bibitem[{Amayuelas et~al.(2024)Amayuelas, Wong, Pan, Chen, and Wang}]{amayuelas2024knowledge}
Alfonso Amayuelas, Kyle Wong, Liangming Pan, Wenhu Chen, and William~Yang Wang. 2024.
\newblock Knowledge of knowledge: Exploring known-unknowns uncertainty with large language models.
\newblock In \emph{Findings of the Association for Computational Linguistics: ACL 2024}, pages 6416--6432.

\bibitem[{Azaria and Mitchell(2023)}]{azaria-mitchell-2023-internal}
Amos Azaria and Tom Mitchell. 2023.
\newblock \href {https://doi.org/10.18653/v1/2023.findings-emnlp.68} {The internal state of an {LLM} knows when it{'}s lying}.
\newblock In \emph{Findings of the Association for Computational Linguistics: EMNLP 2023}, pages 967--976, Singapore. Association for Computational Linguistics.

\bibitem[{Beck et~al.(2016)Beck, Specia, and Cohn}]{beck-etal-2016-exploring}
Daniel Beck, Lucia Specia, and Trevor Cohn. 2016.
\newblock \href {https://doi.org/10.18653/v1/K16-1021} {Exploring prediction uncertainty in machine translation quality estimation}.
\newblock In \emph{Proceedings of the 20th {SIGNLL} Conference on Computational Natural Language Learning}, pages 208--218, Berlin, Germany. Association for Computational Linguistics.

\bibitem[{Beigi et~al.(2024)Beigi, Shen, Yang, Lin, Wang, Mohan, He, Jin, Lu, and Huang}]{beigi-etal-2024-internalinspector}
Mohammad Beigi, Ying Shen, Runing Yang, Zihao Lin, Qifan Wang, Ankith Mohan, Jianfeng He, Ming Jin, Chang-Tien Lu, and Lifu Huang. 2024.
\newblock \href {https://doi.org/10.18653/v1/2024.findings-emnlp.751} {{I}nternal{I}nspector $i^2$: Robust confidence estimation in {LLM}s through internal states}.
\newblock In \emph{Findings of the Association for Computational Linguistics: EMNLP 2024}, pages 12847--12865, Miami, Florida, USA. Association for Computational Linguistics.

\bibitem[{Desai and Durrett(2020)}]{desai-durrett-2020-calibration}
Shrey Desai and Greg Durrett. 2020.
\newblock \href {https://doi.org/10.18653/v1/2020.emnlp-main.21} {Calibration of pre-trained transformers}.
\newblock In \emph{Proceedings of the 2020 Conference on Empirical Methods in Natural Language Processing (EMNLP)}, pages 295--302, Online. Association for Computational Linguistics.

\bibitem[{Dong et~al.(2018)Dong, Quirk, and Lapata}]{dong2018confidence}
Li~Dong, Chris Quirk, and Mirella Lapata. 2018.
\newblock Confidence modeling for neural semantic parsing.
\newblock \emph{arXiv preprint arXiv:1805.04604}.

\bibitem[{Donz{\'e} and Maler(2010)}]{donze2010robust}
Alexandre Donz{\'e} and Oded Maler. 2010.
\newblock Robust satisfaction of temporal logic over real-valued signals.
\newblock In \emph{International conference on formal modeling and analysis of timed systems}, pages 92--106. Springer.

\bibitem[{Duan et~al.(2024)Duan, Cheng, Wang, Zavalny, Wang, Xu, Kailkhura, and Xu}]{duan-etal-2024-shifting}
Jinhao Duan, Hao Cheng, Shiqi Wang, Alex Zavalny, Chenan Wang, Renjing Xu, Bhavya Kailkhura, and Kaidi Xu. 2024.
\newblock \href {https://doi.org/10.18653/v1/2024.acl-long.276} {Shifting attention to relevance: Towards the predictive uncertainty quantification of free-form large language models}.
\newblock In \emph{Proceedings of the 62nd Annual Meeting of the Association for Computational Linguistics (Volume 1: Long Papers)}, pages 5050--5063, Bangkok, Thailand. Association for Computational Linguistics.

\bibitem[{Durrett and Klein(2014)}]{durrett-klein-2014-joint}
Greg Durrett and Dan Klein. 2014.
\newblock \href {https://doi.org/10.1162/tacl_a_00197} {A joint model for entity analysis: Coreference, typing, and linking}.
\newblock \emph{Transactions of the Association for Computational Linguistics}, 2:477--490.

\bibitem[{Farquhar et~al.(2024)Farquhar, Kossen, Kuhn, and Gal}]{farquhar2024detecting}
Sebastian Farquhar, Jannik Kossen, Lorenz Kuhn, and Yarin Gal. 2024.
\newblock Detecting hallucinations in large language models using semantic entropy.
\newblock \emph{Nature}, 630(8017):625--630.

\bibitem[{Garcez et~al.(2019)Garcez, Gori, Lamb, Serafini, Spranger, and Tran}]{garcez2019neural}
Artur~d'Avila Garcez, Marco Gori, Luis~C Lamb, Luciano Serafini, Michael Spranger, and Son~N Tran. 2019.
\newblock Neural-symbolic computing: An effective methodology for principled integration of machine learning and reasoning.
\newblock \emph{arXiv preprint arXiv:1905.06088}.

\bibitem[{Garnelo et~al.(2018)Garnelo, Rosenbaum, Maddison, Ramalho, Saxton, Shanahan, Teh, Rezende, and Eslami}]{garnelo2018conditional}
Marta Garnelo, Dan Rosenbaum, Christopher Maddison, Tiago Ramalho, David Saxton, Murray Shanahan, Yee~Whye Teh, Danilo Rezende, and SM~Ali Eslami. 2018.
\newblock Conditional neural processes.
\newblock In \emph{International conference on machine learning}, pages 1704--1713. PMLR.

\bibitem[{Glenn et~al.(1950)}]{glenn1950verification}
W~Brier Glenn and 1 others. 1950.
\newblock Verification of forecasts expressed in terms of probability.
\newblock \emph{Monthly weather review}, 78(1):1--3.

\bibitem[{Guo et~al.(2017)Guo, Pleiss, Sun, and Weinberger}]{guo2017calibration}
Chuan Guo, Geoff Pleiss, Yu~Sun, and Kilian~Q Weinberger. 2017.
\newblock On calibration of modern neural networks.
\newblock In \emph{International conference on machine learning}, pages 1321--1330. PMLR.

\bibitem[{Ha et~al.(2016)Ha, Dai, and Le}]{ha2016hypernetworks}
David Ha, Andrew Dai, and Quoc~V Le. 2016.
\newblock Hypernetworks.
\newblock \emph{arXiv preprint arXiv:1609.09106}.

\bibitem[{Hou et~al.(2024)Hou, Liu, Qian, Andreas, Chang, and Zhang}]{10.5555/3692070.3692835}
Bairu Hou, Yujian Liu, Kaizhi Qian, Jacob Andreas, Shiyu Chang, and Yang Zhang. 2024.
\newblock Decomposing uncertainty for large language models through input clarification ensembling.
\newblock In \emph{Proceedings of the 41st International Conference on Machine Learning}, ICML'24. JMLR.org.

\bibitem[{Jha et~al.(2019)Jha, Tiwari, Seshia, Sahai, and Shankar}]{jha2019telex}
Susmit Jha, Ashish Tiwari, Sanjit~A Seshia, Tuhin Sahai, and Natarajan Shankar. 2019.
\newblock Telex: learning signal temporal logic from positive examples using tightness metric.
\newblock \emph{Formal Methods in System Design}, 54(3):364--387.

\bibitem[{Jiang et~al.(2021)Jiang, Araki, Ding, and Neubig}]{jiang2021can}
Zhengbao Jiang, Jun Araki, Haibo Ding, and Graham Neubig. 2021.
\newblock How can we know when language models know? on the calibration of language models for question answering.
\newblock \emph{Transactions of the Association for Computational Linguistics}, 9:962--977.

\bibitem[{Kadavath et~al.(2022{\natexlab{a}})Kadavath, Conerly, Askell, Henighan, Drain, Perez, Schiefer, Hatfield-Dodds, DasSarma, Tran-Johnson, Johnston, El-Showk, Jones, Elhage, Hume, Chen, Bai, Bowman, Fort, Ganguli, Hernandez, Jacobson, Kernion, Kravec, Lovitt, Ndousse, Olsson, Ringer, Amodei, Brown, Clark, Joseph, Mann, McCandlish, Olah, and Kaplan}]{kadavath2022languagemodelsmostlyknow}
Saurav Kadavath, Tom Conerly, Amanda Askell, Tom Henighan, Dawn Drain, Ethan Perez, Nicholas Schiefer, Zac Hatfield-Dodds, Nova DasSarma, Eli Tran-Johnson, Scott Johnston, Sheer El-Showk, Andy Jones, Nelson Elhage, Tristan Hume, Anna Chen, Yuntao Bai, Sam Bowman, Stanislav Fort, and 17 others. 2022{\natexlab{a}}.
\newblock \href {https://arxiv.org/abs/2207.05221} {Language models (mostly) know what they know}.
\newblock \emph{Preprint}, arXiv:2207.05221.

\bibitem[{Kadavath et~al.(2022{\natexlab{b}})Kadavath, Conerly, Askell, Henighan, Drain, Perez, Schiefer, Hatfield-Dodds, DasSarma, Tran-Johnson et~al.}]{kadavath2022language}
Saurav Kadavath, Tom Conerly, Amanda Askell, Tom Henighan, Dawn Drain, Ethan Perez, Nicholas Schiefer, Zac Hatfield-Dodds, Nova DasSarma, Eli Tran-Johnson, and 1 others. 2022{\natexlab{b}}.
\newblock Language models (mostly) know what they know.
\newblock \emph{arXiv preprint arXiv:2207.05221}.

\bibitem[{Kamath et~al.(2025)Kamath, Zhang, Xu, Ugare, Singh, and Misailovic}]{kamath2025enforcing}
Adharsh Kamath, Sishen Zhang, Calvin Xu, Shubham Ugare, Gagandeep Singh, and Sasa Misailovic. 2025.
\newblock Enforcing temporal constraints for llm agents.
\newblock \emph{arXiv preprint arXiv:2512.23738}.

\bibitem[{Kendall and Gal(2017)}]{kendall2017uncertainties}
Alex Kendall and Yarin Gal. 2017.
\newblock What uncertainties do we need in bayesian deep learning for computer vision?
\newblock \emph{Advances in neural information processing systems}, 30.

\bibitem[{Kirchhof et~al.(2025)Kirchhof, F{\"u}ger, Golinski, Dhekane, Blaas, and Williamson}]{kirchhof2025self}
Michael Kirchhof, Luca F{\"u}ger, Adam Golinski, Eeshan~Gunesh Dhekane, Arno Blaas, and Sinead Williamson. 2025.
\newblock Self-reflective uncertainties: Do llms know their internal answer distribution?
\newblock In \emph{ICML 2025 Workshop on Reliable and Responsible Foundation Models}.

\bibitem[{Kong et~al.(2014)Kong, Jones, Medina~Ayala, Aydin~Gol, and Belta}]{kong2014temporal}
Zhaodan Kong, Austin Jones, Ana Medina~Ayala, Ebru Aydin~Gol, and Calin Belta. 2014.
\newblock Temporal logic inference for classification and prediction from data.
\newblock In \emph{Proceedings of the 17th international conference on Hybrid systems: computation and control}, pages 273--282.

\bibitem[{Kuhn et~al.(2023)Kuhn, Gal, and Farquhar}]{kuhn2023semantic}
Lorenz Kuhn, Yarin Gal, and Sebastian Farquhar. 2023.
\newblock \href {https://openreview.net/forum?id=VD-AYtP0dve} {Semantic uncertainty: Linguistic invariances for uncertainty estimation in natural language generation}.
\newblock In \emph{The Eleventh International Conference on Learning Representations}.

\bibitem[{Li et~al.(2025{\natexlab{a}})Li, Magesh, and Veeravalli}]{li2025principled}
Jiawei Li, Akshayaa Magesh, and Venugopal~V Veeravalli. 2025{\natexlab{a}}.
\newblock Principled detection of hallucinations in large language models via multiple testing.
\newblock \emph{arXiv preprint arXiv:2508.18473}.

\bibitem[{Li et~al.(2025{\natexlab{b}})Li, Qiang, Moukheiber, and Zhang}]{li2025language}
Yinghao Li, Rushi Qiang, Lama Moukheiber, and Chao Zhang. 2025{\natexlab{b}}.
\newblock \href {https://openreview.net/forum?id=QTrW2HWNXe} {Language model uncertainty quantification with attention chain}.
\newblock In \emph{Second Conference on Language Modeling}.

\bibitem[{Lin et~al.(2022{\natexlab{a}})Lin, Hilton, and Evans}]{lin2022teachingmodelsexpressuncertainty}
Stephanie Lin, Jacob Hilton, and Owain Evans. 2022{\natexlab{a}}.
\newblock \href {https://arxiv.org/abs/2205.14334} {Teaching models to express their uncertainty in words}.
\newblock \emph{Preprint}, arXiv:2205.14334.

\bibitem[{Lin et~al.(2022{\natexlab{b}})Lin, Hilton, and Evans}]{lin2022teaching}
Stephanie Lin, Jacob Hilton, and Owain Evans. 2022{\natexlab{b}}.
\newblock \href {https://openreview.net/forum?id=8s8K2UZGTZ} {Teaching models to express their uncertainty in words}.
\newblock \emph{Transactions on Machine Learning Research}.

\bibitem[{Linard and Tumova(2020)}]{linard2020active}
Alexis Linard and Jana Tumova. 2020.
\newblock Active learning of signal temporal logic specifications.
\newblock In \emph{2020 IEEE 16th International Conference on Automation Science and Engineering (CASE)}, pages 779--785. IEEE.

\bibitem[{Liu et~al.(2025)Liu, Chen, Da, Chen, Lin, and Wei}]{liu2025uncertainty}
Xiaoou Liu, Tiejin Chen, Longchao Da, Chacha Chen, Zhen Lin, and Hua Wei. 2025.
\newblock Uncertainty quantification and confidence calibration in large language models: A survey.
\newblock In \emph{Proceedings of the 31st ACM SIGKDD Conference on Knowledge Discovery and Data Mining V. 2}, pages 6107--6117.

\bibitem[{Lu et~al.(2021)Lu, West, Zellers, Le~Bras, Bhagavatula, and Choi}]{lu-etal-2021-neurologic}
Ximing Lu, Peter West, Rowan Zellers, Ronan Le~Bras, Chandra Bhagavatula, and Yejin Choi. 2021.
\newblock \href {https://doi.org/10.18653/v1/2021.naacl-main.339} {{N}euro{L}ogic decoding: (un)supervised neural text generation with predicate logic constraints}.
\newblock In \emph{Proceedings of the 2021 Conference of the North American Chapter of the Association for Computational Linguistics: Human Language Technologies}, pages 4288--4299, Online. Association for Computational Linguistics.

\bibitem[{Maler and Nickovic(2004)}]{maler2004monitoring}
Oded Maler and Dejan Nickovic. 2004.
\newblock Monitoring temporal properties of continuous signals.
\newblock In \emph{International symposium on formal techniques in real-time and fault-tolerant systems}, pages 152--166. Springer.

\bibitem[{Malinin and Gales(2021)}]{malinin2021uncertainty}
Andrey Malinin and Mark Gales. 2021.
\newblock \href {https://openreview.net/forum?id=jN5y-zb5Q7m} {Uncertainty estimation in autoregressive structured prediction}.
\newblock In \emph{International Conference on Learning Representations}.

\bibitem[{Manakul et~al.(2023)Manakul, Liusie, and Gales}]{manakul2023selfcheckgpt}
Potsawee Manakul, Adian Liusie, and Mark Gales. 2023.
\newblock Selfcheckgpt: Zero-resource black-box hallucination detection for generative large language models.
\newblock In \emph{Proceedings of the 2023 conference on empirical methods in natural language processing}, pages 9004--9017.

\bibitem[{Mao et~al.(2025)Mao, Bisliouk, Nama, and Ruchkin}]{mao-etal-2025-temporalizing}
Zhenjiang Mao, Artem Bisliouk, Rohith Nama, and Ivan Ruchkin. 2025.
\newblock \href {https://doi.org/10.18653/v1/2025.bea-1.65} {Temporalizing confidence: Evaluation of chain-of-thought reasoning with signal temporal logic}.
\newblock In \emph{Proceedings of the 20th Workshop on Innovative Use of NLP for Building Educational Applications (BEA 2025)}, pages 882--890, Vienna, Austria. Association for Computational Linguistics.

\bibitem[{Nicoletti et~al.(2024)Nicoletti, Germiniani, and Pravadelli}]{nicoletti2024mining}
Daniele Nicoletti, Samuele Germiniani, and Graziano Pravadelli. 2024.
\newblock Mining signal temporal logic specifications for hybrid systems.
\newblock In \emph{2024 Forum on Specification \& Design Languages (FDL)}, pages 1--8. IEEE.

\bibitem[{Ott et~al.(2018)Ott, Auli, Grangier, and Ranzato}]{ott2018analyzing}
Myle Ott, Michael Auli, David Grangier, and Marc’Aurelio Ranzato. 2018.
\newblock Analyzing uncertainty in neural machine translation.
\newblock In \emph{International Conference on Machine Learning}, pages 3956--3965. PMLR.

\bibitem[{Pnueli(1977)}]{pnueli1977temporal}
Amir Pnueli. 1977.
\newblock The temporal logic of programs.
\newblock In \emph{18th annual symposium on foundations of computer science (sfcs 1977)}, pages 46--57. ieee.

\bibitem[{Portillo~Wightman et~al.(2023)Portillo~Wightman, Delucia, and Dredze}]{portillo-wightman-etal-2023-strength}
Gwenyth Portillo~Wightman, Alexandra Delucia, and Mark Dredze. 2023.
\newblock \href {https://doi.org/10.18653/v1/2023.trustnlp-1.28} {Strength in numbers: Estimating confidence of large language models by prompt agreement}.
\newblock In \emph{Proceedings of the 3rd Workshop on Trustworthy Natural Language Processing (TrustNLP 2023)}, pages 326--362, Toronto, Canada. Association for Computational Linguistics.

\bibitem[{Raman et~al.(2014)Raman, Maasoumy, and Donz{\'e}}]{raman2014model}
Vasumathi Raman, Mehdi Maasoumy, and Alexandre Donz{\'e}. 2014.
\newblock Model predictive control from signal temporal logic specifications: A case study.
\newblock In \emph{Proceedings of the 4th ACM SIGBED international workshop on design, modeling, and evaluation of cyber-physical systems}, pages 52--55.

\bibitem[{Ribeiro et~al.(2020)Ribeiro, Wu, Guestrin, and Singh}]{ribeiro2020beyond}
Marco~Tulio Ribeiro, Tongshuang Wu, Carlos Guestrin, and Sameer Singh. 2020.
\newblock Beyond accuracy: Behavioral testing of nlp models with checklist.
\newblock \emph{arXiv preprint arXiv:2005.04118}.

\bibitem[{Tian et~al.(2023)Tian, Mitchell, Zhou, Sharma, Rafailov, Yao, Finn, and Manning}]{tian2023justaskcalibrationstrategies}
Katherine Tian, Eric Mitchell, Allan Zhou, Archit Sharma, Rafael Rafailov, Huaxiu Yao, Chelsea Finn, and Christopher~D. Manning. 2023.
\newblock \href {https://arxiv.org/abs/2305.14975} {Just ask for calibration: Strategies for eliciting calibrated confidence scores from language models fine-tuned with human feedback}.
\newblock \emph{Preprint}, arXiv:2305.14975.

\bibitem[{Wang et~al.(2021)Wang, Shi, Wang, Zhang, Zhao, and Zheng}]{wang-etal-2021-beyond-glass}
Ke~Wang, Yangbin Shi, Jiayi Wang, Yuqi Zhang, Yu~Zhao, and Xiaolin Zheng. 2021.
\newblock \href {https://doi.org/10.18653/v1/2021.findings-emnlp.401} {Beyond glass-box features: Uncertainty quantification enhanced quality estimation for neural machine translation}.
\newblock In \emph{Findings of the Association for Computational Linguistics: EMNLP 2021}, pages 4687--4698, Punta Cana, Dominican Republic. Association for Computational Linguistics.

\bibitem[{Wang et~al.(2022)Wang, Wei, Schuurmans, Le, Chi, Narang, Chowdhery, and Zhou}]{wang2022self}
Xuezhi Wang, Jason Wei, Dale Schuurmans, Quoc Le, Ed~Chi, Sharan Narang, Aakanksha Chowdhery, and Denny Zhou. 2022.
\newblock Self-consistency improves chain of thought reasoning in language models.
\newblock \emph{arXiv preprint arXiv:2203.11171}.

\bibitem[{Wang et~al.(2023)Wang, Wei, Schuurmans, Le, Chi, Narang, Chowdhery, and Zhou}]{wang2023selfconsistency}
Xuezhi Wang, Jason Wei, Dale Schuurmans, Quoc~V Le, Ed~H. Chi, Sharan Narang, Aakanksha Chowdhery, and Denny Zhou. 2023.
\newblock \href {https://openreview.net/forum?id=1PL1NIMMrw} {Self-consistency improves chain of thought reasoning in language models}.
\newblock In \emph{The Eleventh International Conference on Learning Representations}.

\bibitem[{Wei et~al.(2022)Wei, Wang, Schuurmans, Bosma, Xia, Chi, Le, Zhou et~al.}]{wei2022chain}
Jason Wei, Xuezhi Wang, Dale Schuurmans, Maarten Bosma, Fei Xia, Ed~Chi, Quoc~V Le, Denny Zhou, and 1 others. 2022.
\newblock Chain-of-thought prompting elicits reasoning in large language models.
\newblock \emph{Advances in neural information processing systems}, 35:24824--24837.

\bibitem[{Xiong et~al.(2024)Xiong, Hu, Lu, LI, Fu, He, and Hooi}]{xiong2024can}
Miao Xiong, Zhiyuan Hu, Xinyang Lu, YIFEI LI, Jie Fu, Junxian He, and Bryan Hooi. 2024.
\newblock \href {https://openreview.net/forum?id=gjeQKFxFpZ} {Can {LLM}s express their uncertainty? an empirical evaluation of confidence elicitation in {LLM}s}.
\newblock In \emph{The Twelfth International Conference on Learning Representations}.

\bibitem[{Yao et~al.(2023)Yao, Yu, Zhao, Shafran, Griffiths, Cao, and Narasimhan}]{yao2023tree}
Shunyu Yao, Dian Yu, Jeffrey Zhao, Izhak Shafran, Tom Griffiths, Yuan Cao, and Karthik Narasimhan. 2023.
\newblock Tree of thoughts: Deliberate problem solving with large language models.
\newblock \emph{Advances in neural information processing systems}, 36:11809--11822.

\bibitem[{Yu et~al.(2023)Yu, Lin, Leung, Ar\'{e}chiga, and Pavone}]{10.1177/02783649221082115}
Jingjin Yu, Ming~C. Lin, Karen Leung, Nikos Ar\'{e}chiga, and Marco Pavone. 2023.
\newblock \href {https://doi.org/10.1177/02783649221082115} {Backpropagation through signal temporal logic specifications: Infusing logical structure into gradient-based methods}.
\newblock \emph{Int. J. Rob. Res.}, 42(6):356–370.

\end{thebibliography}
\end{document}